\documentclass[10pt,twocolumn,letterpaper]{article}
\usepackage[pagenumbers]{cvpr}
\usepackage{algorithm}
\usepackage{algpseudocode}
\usepackage{multirow}
\usepackage[percent]{overpic}
\usepackage{trimclip}

\usepackage{amsthm}
\newtheorem{assumption}{Assumption}
\AtEndPreamble{\crefname{assumption}{Assumption}{Assumptions}%
  \Crefname{assumption}{Assumption}{Assumptions}}

\usepackage{microtype}

\definecolor{cvprblue}{rgb}{0.21,0.49,0.74}
\usepackage[breaklinks,colorlinks,allcolors=cvprblue]{hyperref}
\title{KLI-Quant: Task-Sensitive W4A4 Quantization for Diffusion Models via Decoupled Activation and Weight Error Control}
\author{%
  Mo Song\textsuperscript{1,*}\quad
  Sen Dong\textsuperscript{2,*}\quad
  Ruiliang Zhou\textsuperscript{2,*}\quad
  Zhijie Yang\textsuperscript{3}\quad
  Yan Wang\textsuperscript{4}\quad
  Yuanzhe Li\textsuperscript{5}\\
  Qinqin Chen\quad
  Jingyu Xin\textsuperscript{6}\quad
  Kang He\textsuperscript{2}\quad
  Chengru Song\textsuperscript{2,\textdagger}\quad
  Hayden Kwok-Hay So\textsuperscript{1}\\[3pt]
  {\small \textsuperscript{1}The University of Hong Kong\quad \textsuperscript{2}KlingAI Research\quad \textsuperscript{3}Peking University}\\
  {\small \textsuperscript{4}Shanghai Jiao Tong University\quad \textsuperscript{5}Dalian University of Technology\quad \textsuperscript{6}University of Arizona}\\
  {\small \textsuperscript{*}These authors contributed equally and share first authorship.\quad \textsuperscript{\textdagger}Corresponding author.}%
}
\hypersetup{pdftitle={\thetitle},
  pdfauthor={Mo Song, Sen Dong, Ruiliang Zhou, Zhijie Yang, Yan Wang, Yuanzhe Li, Qinqin Chen, Jingyu Xin, Kang He, Chengru Song, Hayden Kwok-Hay So}}

\makeatletter
\renewcommand{\@maketitle}{%
  \begingroup
  \centering
  \vspace*{20pt}
  {\Large\bfseries \@title\par}
  \vspace{10pt}
  {\large\begin{tabular}{@{}c@{}}\@author\end{tabular}\par}
  \endgroup
  \vspace{10pt}%
}
\makeatother

\begin{document}
\twocolumn[{%
\renewcommand\twocolumn[1][]{#1}%
\maketitle
\noindent\begin{minipage}{\textwidth}
  \centering
  \begingroup%
\def\qfont{\fontfamily{phv}\fontseries{m}\fontshape{n}\fontsize{9}{10.2}\selectfont}%
\noindent\hspace*{58pt}%
\begin{overpic}[width=\dimexpr\linewidth-61pt\relax,trim=30.72bp 7.8bp 0bp 0bp,clip]{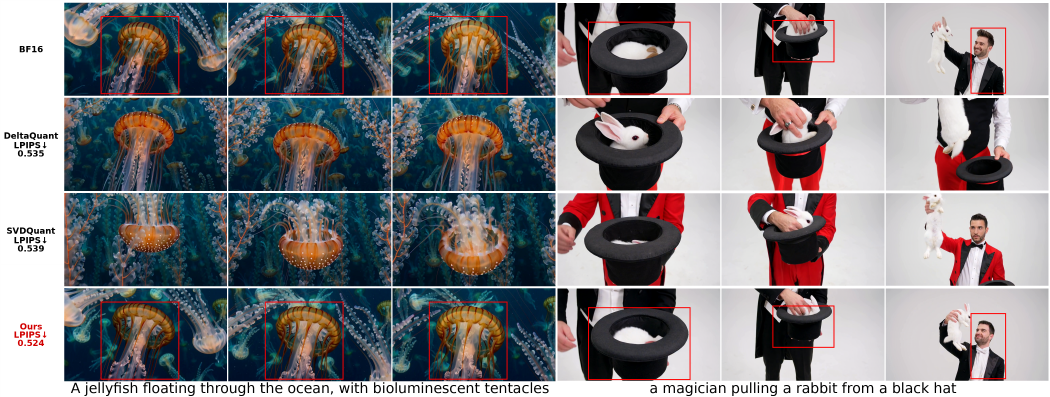}%
  \put(0,33.582657){\makebox(0,0)[r]{\makebox[58pt][c]{\qfont\shortstack[c]{BF16}}}}%
  \put(0,23.947769){\makebox(0,0)[r]{\makebox[58pt][c]{\qfont\shortstack[c]{DeltaQuant\\LPIPS$\downarrow$\\0.535}}}}%
  \put(0,14.312880){\makebox(0,0)[r]{\makebox[58pt][c]{\qfont\shortstack[c]{SVDQ\\LPIPS$\downarrow$\\0.539}}}}%
  \put(0,4.677992){\makebox(0,0)[r]{\makebox[58pt][c]{\color{red}\qfont\shortstack[c]{Ours\\LPIPS$\downarrow$\\0.524}}}}%
\end{overpic}\par%
\noindent\hspace*{58pt}\begin{minipage}{\dimexpr\linewidth-61pt\relax}%
\qfont\noindent%
\parbox[t]{0.495\linewidth}{\centering Bioluminescent jellyfish floating through the ocean}\hfill%
\parbox[t]{0.495\linewidth}{\centering Magician pulling a rabbit from a black hat}%
\end{minipage}\par%
\endgroup\ignorespaces%

  \captionsetup{hypcap=false}
  \captionof{figure}{Four-step video generation on MiniMax-H3 Fast. KLI-Quant better preserves BF16 composition and details.}
  \label{fig:h3-dist-main-qualitative}
\end{minipage}
\par\vspace{0.5em}

}]
\newskip\kliqfs \newskip\kliqtfs \newskip\kliqdfs \newskip\kliqdtfs
\kliqfs=\floatsep \kliqtfs=\textfloatsep
\kliqdfs=\dblfloatsep \kliqdtfs=\dbltextfloatsep
\setlength{\floatsep}{6pt plus 2pt minus 2pt}
\setlength{\textfloatsep}{8pt plus 2pt minus 2pt}
\setlength{\dblfloatsep}{6pt plus 2pt minus 2pt}
\setlength{\dbltextfloatsep}{8pt plus 2pt minus 2pt}
\setcounter{dbltopnumber}{3}
\begin{abstract}
Diffusion models have become the leading approach to high-fidelity image and
video generation, but their large memory footprint and long sampling latency
make deployment costly. Quantizing both weights and activations to 4 bits
(W4A4) substantially reduces both, but makes generation quality hard to
preserve. At this aggressive precision, keeping a small high-precision
low-rank branch to absorb outliers, as in SVDQ, has proven effective.
Such methods, however, still fall short for two reasons: their calibration
objective, the layer output error, fails to reflect the error in the
network output of the denoising step (the step output error), and their
bypass allocation primarily targets weight reconstruction rather than
transmitted activation error. We
propose KLI-Quant, a W4A4 post-training quantization framework for diffusion
models that addresses both limitations. First, KLI-Quant weights the
calibration objective by downstream sensitivity, estimated from a frozen
full-precision model, so that the objective tracks the step output error
rather than the layer output error. Second, it retains this low-rank bypass
but regroups the deployed layer error into an activation term that depends
on the bypass but not on residual quantization codes, and a weight term
measured under quantized inputs. This grouping defines complementary
targets for two-stage calibration, first allocating the bypass against
activation error and then calibrating the residual weights under quantized
inputs. We further provide efficient W4A4 kernels for the resulting
two-branch layer. On MiniMax-H3, KLI-Quant achieves state-of-the-art
video generation quality among the evaluated W4A4 methods, while
achieving a $3.12\times$ average GEMM speedup and a $1.93\times$ end-to-end
speedup over BF16.
\end{abstract}

\section{Introduction}
\label{sec:intro}

Diffusion models~\cite{ho2020ddpm,rombach2022ldm} with transformer backbones
(DiTs)~\cite{peebles2023dit} support high-quality image and video
generation~\cite{esser2024sd3,bfl2024flux,kong2024hunyuanvideo,wan2025wan,minimax2026h3},
but repeatedly evaluating large networks makes inference costly.
Quantizing both weights and activations to 4 bits reduces storage and
enables low-bit matrix multiplication. The difficulty is preserving the
reference computation when both operands are rounded to a sparse numerical
grid. Errors introduced throughout the denoiser can alter scene structure,
fine details, and temporal consistency.

High-precision low-rank bypasses offer a practical way to retain selected
weight components while quantizing the remaining residual.
SVDQ~\cite{li2025svdquant} introduces this layout for diffusion models,
and DeltaQuant~\cite{li2026deltaquant} and
H-SVDQ~\cite{wang2026hsvdquant} build on it. However, allocating the
bypass primarily for weight reconstruction does not directly target the
activation error transmitted through the residual. Our measurements show
that activation error remains substantial even after residual weight
reconstruction improves (\cref{sec:motivation-activation-weight}). The
bypass receives unquantized inputs, so each weight component assigned to it
avoids direct exposure to activation quantization. Its limited capacity
should therefore protect components that transmit consequential activation
errors, rather than only those with large weight magnitude.

Determining which errors are consequential requires more than a local
reconstruction metric. Layer output errors of equal magnitude can produce
different errors at the final output of a denoising step, depending on their
directions and propagation through subsequent layers
(\cref{sec:motivation-local-global}). Reducing local error alone can thus
spend precision on directions with limited downstream influence.
Activation-error statistics identify where quantization is inaccurate,
while downstream sensitivity identifies where that inaccuracy matters for the
prediction. Together, they provide a criterion for allocating high-precision
capacity according to the error it protects against.

\begin{figure*}[t]
  \centering
  \includegraphics[width=0.90\textwidth,pagebox=cropbox,clip]{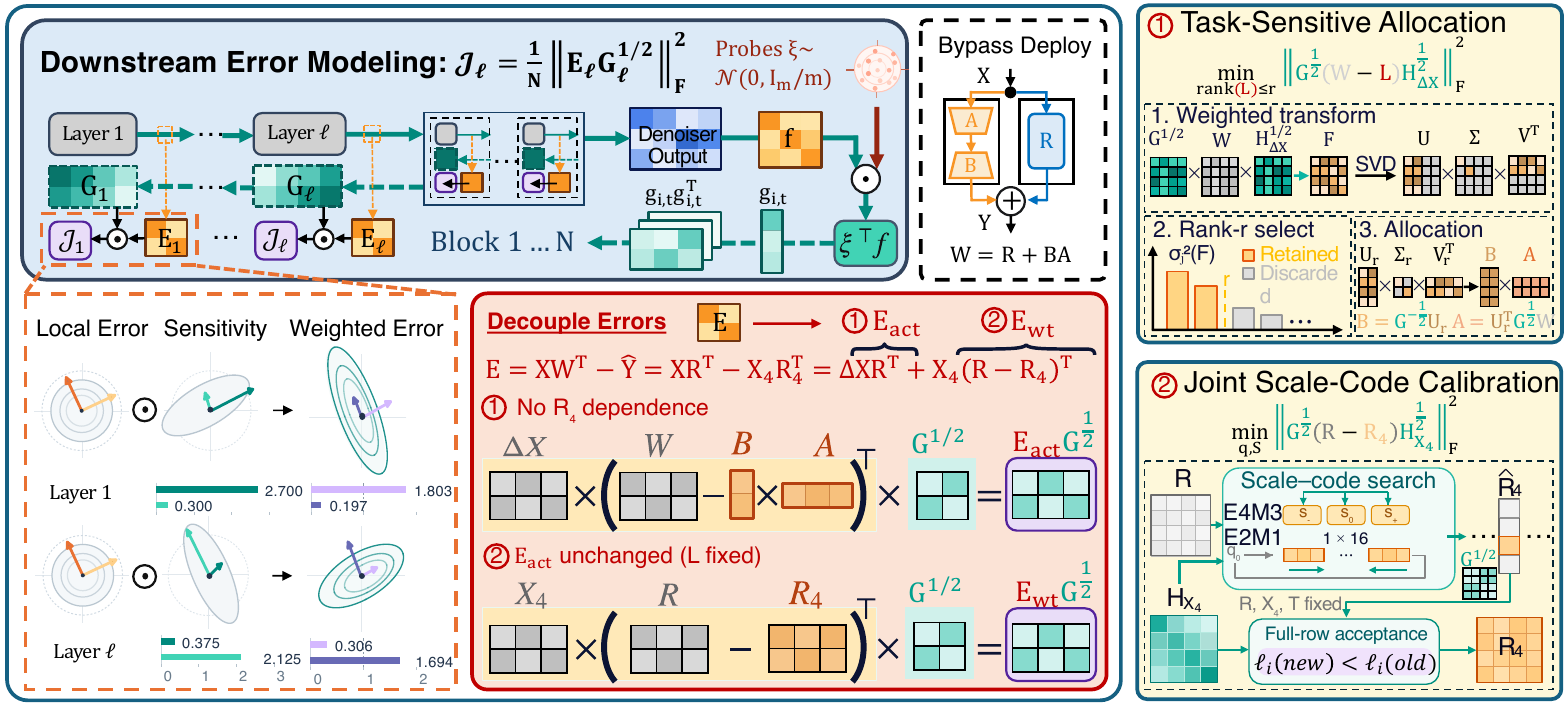}
  \caption{Overview of KLI-Quant. Left: objective modeling and error
    decomposition. Right: the two-stage method.}
  \label{fig:overview}
\end{figure*}

We propose KLI-Quant, a post-training quantization framework that
assigns distinct roles to activation-error protection and residual weight
reconstruction (\cref{fig:overview}). Our error model separates the layer
discrepancy into the activation error transmitted by the residual before
weight quantization and the weight-reconstruction error evaluated under
quantized inputs. The first contribution changes with bypass allocation but
not with the residual codes. Once the bypass is fixed, the second gives a
reconstruction target matched to the deployed residual branch. A shared
downstream sensitivity metric connects both contributions to predicted
denoising-step output error. This separation defines complementary
optimization targets without requiring the two errors to be independent.

KLI-Quant implements these targets in two stages. First, measured activation-error
second moments and downstream sensitivity estimated from the frozen reference model
guide a weighted low-rank allocation. After storing the bypass factors, we
form the residual from those actual factors and keep the bypass fixed.
Second, joint scale--code (JSC) calibration searches representable block
scales and refines the associated 4-bit codes under the quantized-input
distribution and the same downstream metric. Scales and codes must be
considered together because each scale sets the grid for an entire block.
Input correlations and downstream sensitivity also couple reconstruction
decisions. JSC accounts for these dependencies through compensated
construction, conditional refinement, and acceptance against the coupled
objective. For deployment, a fused kernel accumulates the low-bit residual
and BF16 rank-up correction in shared FP32 registers, avoiding an
intermediate output write and reload.

Experiments compare methods adapted to the same NVFP4 format across image
and video generators. KLI-Quant lowers LPIPS by 4.7--10.8\% relative to the
strongest compared baseline on HunyuanVideo and Wan2.2, and achieves the best
PSNR and LPIPS among quantized methods in all six Qwen-Image
dataset/checkpoint settings. On MiniMax-H3 Fast, activation-term bypass
allocation reduces four-step mean output MSE by 20.39\% relative to
weight-SVD allocation under the same $G$-weighted JSC residual calibration. The fused kernel achieves a 3.12-fold mean GEMM speedup over BF16
(\cref{sec:experiments,app:latency-details}).

\noindent Our contributions are as follows.
\begin{itemize}\itemsep0pt
  \item An error-control formulation that assigns high-precision capacity to
        transmitted activation error, using measured activation-error
        statistics and frozen-reference downstream sensitivity
        (\cref{sec:error-modeling,sec:stage-allocation}).
  \item JSC residual calibration that jointly selects representable scales
        and codes under deployed quantized inputs, retaining input
        correlations and full downstream sensitivity
        (\cref{sec:stage-calibration}).
  \item A fused implementation of the two-branch layer and evaluations of
        generation quality, reference-output fidelity, allocation
        efficiency, and execution latency
        (\cref{sec:kernel,sec:experiments}).
\end{itemize}

\section{Related work}
\label{sec:related-work}

\subsection{Compensation \& downstream sensitivity}

Compensation preserves local outputs, while sensitivity weights downstream
effects. GPTQ~\cite{frantar2023gptq} compensates sequential weight rounding
using input-derived second-order information. GPTAQ~\cite{li2025gptaq} includes activation quantization and
upstream drift. BRECQ~\cite{li2021brecq} and
PD-Quant~\cite{liu2023pdquant} use diagonal Fisher-weighted block
reconstruction and prediction differences, respectively.
IO-SVD~\cite{abbasi2026iosvd} combines input whitening with predictive-KL
output weighting. Align, Then Correct~\cite{song2026align} combines
Fisher-weighted asymmetric low-rank alignment with remeasured
natural-gradient correction of frozen quantized weights.

Relative to GPTAQ, KLI-Quant allocates high-precision capacity to protect against
activation error before residual compensation. Unlike IO-SVD, it targets
activation error transmitted through a retained residual rather than
dense-operator replacement. Unlike Align, Then Correct,
it allocates the bypass before residual quantization and fixes it during
calibration under the shared downstream metric.

\subsection{High-precision low-rank bypass}

Bypass-free approaches use outlier rescaling
(SmoothQuant~\cite{xiao2023smoothquant}), timestep-aware calibration
(Q-Diffusion~\cite{li2023qdiffusion}), dynamic mixed-precision quantization
(ViDiT-Q~\cite{zhao2025viditq}), rotated auto-scaling with mixed precision
(DVD-Quant~\cite{li2026dvdquant}), or salient
data and sparse-token distillation
(S2Q-VDiT~\cite{feng2025s2qvdit}). Low-rank corrections
reconstruct weight error by truncated SVD (LoRC~\cite{yao2024lorc}),
incorporate activation magnitudes (the $\mathrm{L}^{2}$QER variant of
LQER~\cite{zhang2024lqer}) or input autocorrelations
(QERA~\cite{zhang2025qera}), or jointly fit quantized weights and the
bypass (LRC~\cite{scetbon2024lrc}).

SVDQ~\cite{li2025svdquant} uses weight-SVD allocation followed by
GPTQ residual calibration. DeltaQuant~\cite{li2026deltaquant} adds FP8 core
tokens and 4-bit spatiotemporal differences. H-SVDQ~\cite{wang2026hsvdquant}
combines input-weighted SVD, Hessian deflation, bypass refitting, and diagonal
grouped-range activation-noise modeling. KLI-Quant allocates the full weight
using measured activation-error moments and downstream sensitivity,
retaining correlations and bias. Unlike QERA, it allocates before
quantization; unlike LRC and H-SVDQ, it fixes the bypass during residual
calibration.
ARHQ~\cite{wang2026arhq} uses an activation-only allocation objective,
with ARHQ-w further incorporating weight error. We derive complementary
allocation and quantized-input residual calibration targets from the
deployed W4A4 error, with measured interactions and residual-cost variation
motivating their separation.

\subsection{Adaptive rounding and scale selection}

Rounding and scales jointly determine output reconstruction.
On a fixed grid, nearest rounding minimizes scalar error, not necessarily
output error, because inputs combine individual errors.
AdaRound~\cite{nagel2020adaround} learns relaxed rounding, while
FlexRound~\cite{lee2023flexround} learns division-based scaling. For NVFP4,
Four Over Six~\cite{cook2025fouroversix} maps block maxima toward four or six.
ScaleSearch~\cite{gupta2026scalesearch} scores neighboring E4M3 scales with
nearest-rounded E2M1 codes by unweighted block error,
H-Scale~\cite{yu2026hscale} adds diagonal input
weighting, and OSFP4~\cite{bendavid2026osfp4} optimizes smoothing and scales
through a multiplicative-dither model before discrete scale selection.
KLI-Quant searches representable scales and
conditionally refines codes without learned auxiliary rounding parameters. Full
quantized-input correlations and downstream sensitivity guide grouped proposals
with exact coupled acceptance, beyond unweighted or diagonal scale scoring.

\section{Preliminaries and observations}
\label{sec:background}
\label{sec:preliminary}

\subsection{Preliminaries}
\label{sec:prelim-format}

\textbf{NVFP4 representation.} NVFP4 stores values as 4-bit E2M1 codes with
an FP32 tensor scale $T$ and an E4M3 scale $S_b$ shared by each block of 16
consecutive elements~\cite{alvarez2025nvfp4}. For a block $V_b$ of a tensor
$V$ with positive scale, direct rounding and reconstruction are
\begin{equation}
  q_b=\Pi_{\mathrm{E2M1}}\!\left(\frac{V_b}{TS_b}\right),\qquad
  Q(V)_b=TS_bq_b,
  \label{eq:motivation-nvfp4}
\end{equation}
where $\Pi_{\mathrm{E2M1}}$ rounds to the nearest E2M1 value (maximum
magnitude $6$). In W4A4, weights are quantized offline and activations at
inference.

\textbf{Low-rank bypass and quantized residual.}
SVDQ~\cite{li2025svdquant} splits a layer weight $W$ into a
high-precision rank-$r$ bypass $L=BA$ and a residual $R=W-L$ quantized to
$R_4$. With $N$ input tokens in $X$, the deployed layer computes
\begin{equation}
  \widehat Y=XL^\top+X_4R_4^\top,
  \label{eq:prelim-deployed}
\end{equation}
where $X_4=Q_{\mathrm{act}}(X)$. Only the residual branch runs in W4A4.

\subsection{Observation 1: error direction matters}
\label{sec:motivation-local-global}

\begin{figure}[t]
  \centering
  \captionsetup[subfigure]{justification=centering}
  \begin{subfigure}[t]{0.48\linewidth}
    \centering
    \includegraphics[width=\linewidth]{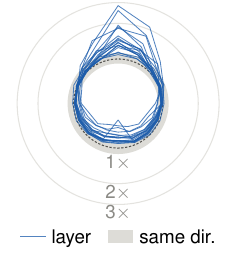}
    \caption{Step output error per direction}
    \label{fig:direction}
  \end{subfigure}\hfill
  \begin{subfigure}[t]{0.48\linewidth}
    \centering
    \includegraphics[width=\linewidth]{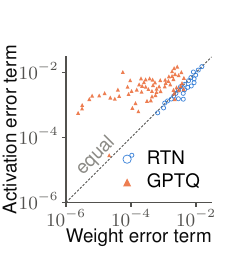}
    \caption{Activation vs.\ weight error}
    \label{fig:act-weight}
  \end{subfigure}
  \vspace{-0.5em}
  \caption{Two observations on MiniMax-H3 Fast.}
  \label{fig:observations}
\end{figure}

Reconstruction-based calibration treats layer output errors of equal size as
equally costly. On the four-step MiniMax-H3 Fast~\cite{minimax2026h3}, we split the output space of a layer into 16 orthogonal
subspaces and inject a Gaussian error of the same energy into each in turn,
then measure the step output error, the MSE of the step's output relative to
BF16 (18 layers, 4 prompts, all 4 steps; \cref{app:observations}). In
\cref{fig:direction}, the most and least harmful directions of a layer differ
by up to $4.4\times$ (median $1.75\times$), against at most $1.2\times$ when one
direction is repeated with new random values. An unweighted reconstruction
loss cannot separate these directions, which motivates weighting the layer
objective by the step output error (\cref{sec:error-modeling}).

\subsection{Observation 2: activation error persists}
\label{sec:motivation-activation-weight}

SVDQ~\cite{li2025svdquant} bounds the layer output error
$E=XW^\top-\widehat Y$ using $\Delta_X=X-X_4$ and $\Delta_R=R-R_4$:
\begin{equation}
  \|E\|_F=\|X\Delta_R^\top+\Delta_XR_4^\top\|_F
  \le\|X\Delta_R^\top\|_F+\|\Delta_XR_4^\top\|_F,
  \label{eq:svd-error-bound}
\end{equation}
and allocates a weight-SVD bypass after smoothing~\cite{xiao2023smoothquant}.
On rank-64 SVDQ (24 held-out prompts; \cref{app:observations}), we
measure activation-only error $\Delta_XR^\top$ and weight-only error
$X\Delta_R^\top$ by quantizing only the residual-branch input or residual,
respectively. The errors are comparable with RTN (median ratio $1.0$,
\cref{fig:act-weight}). GPTQ~\cite{frantar2023gptq} reduces weight-only
error by a median $4.9\times$, while activation-only error is unchanged
and larger in 97\% of the evaluated layers. The unquantized bypass can
reduce this activation error, motivating activation-aware allocation
(\cref{sec:error-modeling}).

\section{Method}
\label{sec:method}

\subsection{Downstream-weighted error decomposition}
\label{sec:error-modeling}
\label{sec:method-overview}

For a layer of the form \cref{eq:prelim-deployed} with weight
$W\in\mathbb R^{o\times d}$, calibration inputs $X\in\mathbb R^{N\times d}$,
and a bypass $L=BA$ of fixed rank $r$ ($A\in\mathbb R^{r\times d}$,
$B\in\mathbb R^{o\times r}$), the layer output error is
$E=XW^\top-\widehat Y\in\mathbb R^{N\times o}$.

\textbf{Downstream weighting.} Equal layer output errors in different output
directions cause unequal step output errors
(\cref{sec:motivation-local-global}), so we weight $E$ by its effect on the
step output error. Let $f\in\mathbb R^m$ be the output of the frozen
reference model at the current denoising step and $\delta f$ its change
caused by $E$, so that the step output error is $\|\delta f\|_2^2/m$. Let
$e_t^\top$ be row $t$ of $E$, $J_t$ the
Jacobian from the layer output of token $t$ to $f$, and
$G_t=J_t^\top J_t/m$. First-order linearization, omission of cross-token and
cross-layer interactions, and replacing $G_t$ by its average $\bar G$ over
calibration inputs and tokens give (\cref{app:sensitivity})
\begin{equation}
  \frac{\|\delta f\|_2^2}{m}
   \approx\sum_te_t^\top G_te_t
   \approx\sum_te_t^\top\bar Ge_t,
  \label{eq:step-error-chain}
\end{equation}
with $t$ over the tokens of the step. We estimate $\bar G$ with Gaussian
probes at the reference output: for $\xi\sim\mathcal N(0,I_m/m)$, the
vector--Jacobian product $g_t=J_t^\top\xi$ satisfies
$\mathbb E_\xi[g_tg_t^\top]=G_t$, so averaging $g_tg_t^\top$ over
calibration inputs, tokens, and probes estimates $\bar G$ without
bias. We normalize per layer to
$G=(\bar G/\alpha+\rho I_o)/(1+\rho)$ with
$\alpha=\operatorname{tr}(\bar G)/o$ and $\rho=10^{-4}$, so that $G\succ0$.
Summing \cref{eq:step-error-chain} over the calibration inputs and dividing
by their total token count $N$ gives the surrogate
\begin{equation}
  \mathcal J=\frac1N\sum_te_t^\top Ge_t=\frac1N\|EG^{1/2}\|_F^2.
  \label{eq:downstream-surrogate}
\end{equation}

\textbf{Error decomposition.} The bypass receives the unquantized input, so
$XL^\top$ cancels in $E$. Adding and subtracting $X_4R^\top$ splits the
remainder exactly into an activation term and a weight term,
\begin{equation}
  E=XR^\top-X_4R_4^\top
   =\underbrace{\Delta_XR^\top}_{E_{\mathrm{act}}}
    +\underbrace{X_4\Delta_R^\top}_{E_{\mathrm{wt}}}.
  \label{eq:error-decomposition}
\end{equation}
The activation term depends on the bypass through $R=W-L$ and not on $R_4$.
The weight term is the residual error under the input $X_4$ that the
residual branch receives. In the grouping of \cref{eq:svd-error-bound}, by
contrast, the activation term $\Delta_XR_4^\top$ changes with $R_4$ during
residual calibration (\cref{app:decomposition}).

Substituting \cref{eq:error-decomposition} into
\cref{eq:downstream-surrogate} and expanding the square gives
\begin{equation}
  \begin{aligned}
    \mathcal J&=\mathcal J_{\mathrm{act}}(L)
                +\mathcal J_{\mathrm{wt}}(L,R_4)+\mathcal J_\times,\\
    \mathcal J_{\mathrm{act}}(L)
      &=\|G^{1/2}(W-L)H_{\Delta_X}^{1/2}\|_F^2,\\
    \mathcal J_{\mathrm{wt}}(L,R_4)
      &=\|G^{1/2}(R-R_4)H_{X_4}^{1/2}\|_F^2,
  \end{aligned}
  \label{eq:three-term-identity}
\end{equation}
where, as in GPTQ~\cite{frantar2023gptq}, each term is weighted by the
Hessian of its input, $H_{\Delta_X}=\Delta_X^\top\Delta_X/N$ and
$H_{X_4}=X_4^\top X_4/N$. The signed cross term $\mathcal J_\times$ is
small: on MiniMax-H3 Fast it is $-0.78\%$ of $\mathcal J$ and negative in 309 of the
312 linear layers (\cref{app:interactions}). Dropping it gives
the calibration objective
\begin{equation}
  \boxed{\min_{\substack{\operatorname{rank}(L)\le r\\
                        R_4\in\mathcal Q_{\mathrm{W4}}(L)}}
    \mathcal J_{\mathrm{act}}(L)+\mathcal J_{\mathrm{wt}}(L,R_4),}
  \label{eq:two-component-objective}
\end{equation}
where $\mathcal Q_{\mathrm{W4}}(L)$ is the set of NVFP4 reconstructions of
$R=W-L$ at a fixed tensor scale.

\subsection{Two-stage optimization}
\label{sec:allocation-calibration}

The two terms of \cref{eq:two-component-objective} remain coupled through
$R=W-L$. We separate them with the following assumption.

\begin{assumption}[Bypass-dominated activation term]
  \label{asm:floor}
  Across bypasses $L$ of the same rank, the attainable weight cost
  $\Phi_{\mathrm W}(L)=\min_{R_4\in\mathcal Q_{\mathrm{W4}}(L)}
    \mathcal J_{\mathrm{wt}}(L,R_4)$ changes much less than
  $\mathcal J_{\mathrm{act}}(L)$.
\end{assumption}

A low-rank bypass does not generally capture the element-wise NVFP4 rounding
pattern of the residual, so it changes the weight term mainly through the
residual energy. Across the rank-64 bypasses evaluated on MiniMax-H3 Fast,
the achieved weight cost varies a median $0.05$ times as much as the
activation cost (\cref{sec:bypass-choice}). This observation
motivates the two-stage optimization: Stage~1 constructs $L=BA$
for $\mathcal J_{\mathrm{act}}$, and Stage~2 forms $R=W-BA$ from the stored
factors and calibrates $R_4$ for $\mathcal J_{\mathrm{wt}}$
(\cref{app:floor}).

\subsubsection{Stage 1: task-sensitive allocation}
\label{sec:stage-allocation}

Stage~1 solves $\min_{\operatorname{rank}(L)\le r}
\|G^{1/2}(W-L)H_{\Delta_X}^{1/2}\|_F^2$. The right metric prioritizes input
directions where activation quantization is inaccurate, and the left metric
weights their downstream consequences. Define the linear map
$F(M)=G^{1/2}MH_{\Delta_X}^{1/2}$, which weights a matrix by both metrics.
Because $F$ is linear, $F(W-L)=F(W)-F(L)$ and
\begin{equation}
  \mathcal J_{\mathrm{act}}(L)=\|F(W)-F(L)\|_F^2.
  \label{eq:allocation-objective}
\end{equation}
Here $F(W)$ is fixed, and $F(L)$ has rank at most $r$ because $L$ does, so
$\mathcal J_{\mathrm{act}}(L)$ is the error of a rank-$r$ approximation of
$F(W)$. Let $F(W)=\sum_k\sigma_ku_kv_k^\top$ be its singular value
decomposition with $\sigma_1\ge\sigma_2\ge\cdots$, and
$U_r=[u_1,\ldots,u_r]$. By the Eckart--Young theorem,
$\mathcal J_{\mathrm{act}}(L)\ge\sum_{k>r}\sigma_k^2$, with equality when
$F(L)$ keeps the $r$ largest singular values of $F(W)$ and their singular
vectors,
\begin{equation}
  F(L)=\sum\nolimits_{k=1}^r\sigma_ku_kv_k^\top=U_rU_r^\top F(W).
  \label{eq:allocation-target}
\end{equation}

For exact leading singular vectors $U_r$, choose
\begin{equation}
  B=G^{-1/2}U_r,\qquad A=U_r^\top G^{1/2}W.
  \label{eq:bypass-factors}
\end{equation}
In exact arithmetic, these factors attain the bound and give a global
minimizer (\cref{app:allocation}). In practice, we approximate $U_r$ per
layer by randomized subspace iteration with bounded Rayleigh--Ritz
enhancement and store $A,B$ in deployment precision, so the
exact minimum is not guaranteed. We recompute $R=W-BA$ from the stored
factors. Only $G\succ0$ is inverted, so $H_{\Delta_X}$ may be singular.
The metrics are absorbed into $A,B$ and add no inference cost.

\subsubsection{Stage 2: joint scale--code calibration}
\label{sec:stage-calibration}

\begin{algorithm}[t]
  \caption{Joint scale--code (JSC) calibration}
  \label{alg:joint-calibration}
  \small
  \begin{algorithmic}[1]
    \Require $R$, $G$, $H_{X_4}$, group size $n_{\mathrm g}$,
             tolerance $\tau$
    \State $S\gets$ absmax block scales; $q\gets$ nearest codes
           \label{alg:jsc-init}
    \Repeat \Comment{pass $k=0,1,\ldots$}
      \For{each group $\mathcal I$ of $n_{\mathrm g}$ rows, in sequence}
        \For{each row $i\in\mathcal I$, in parallel}
          \State $\tilde R_i\gets$ shifted target by \cref{eq:shifted-target}
                 \label{alg:jsc-target}
          \For{each block $b$ of row $i$, left to right}
            \For{scale $s$ within 2 E4M3 steps of $S_{i,b}$}
              \State round on grid $s$, compensated if $k=0$
                     \label{alg:jsc-round}
              \State refine its codes by coordinate descent
                     \label{alg:jsc-refine}
            \EndFor
            \State keep the $(s,q_{i,b})$ with the lowest cost
                   \label{alg:jsc-keep}
            \State \textbf{if} $k=0$: correct later columns by
                   \cref{eq:construction-feedback}
                   \label{alg:jsc-feedback}
          \EndFor
        \EndFor
        \State accept rows greedily by exact
               $\Delta\mathcal J_{\mathrm{wt}}<0$
               \label{alg:jsc-accept}
      \EndFor
    \Until{relative decrease of $\mathcal J_{\mathrm{wt}}$ per pass
           $<\tau$}
           \label{alg:jsc-stop}
  \end{algorithmic}
\end{algorithm}

With $A,B$ fixed, Stage~2 quantizes $R=W-BA$ in the format of
\cref{eq:motivation-nvfp4} with $T$ fixed. Each block scale $S_{i,b}$ sets
the grid of its 16 codes $q_{i,b}$, so JSC calibration optimizes both,
\begin{equation}
  \min_{q,S}\ \mathcal J_{\mathrm{wt}}
   =\bigl\|G^{1/2}(R-R_4)H_{X_4}^{1/2}\bigr\|_F^2.
  \label{eq:code-scale-problem}
\end{equation}

\textbf{Compensated rounding.} Because $H_{X_4}$ couples the input columns,
nearest rounding does not minimize the weighted error of a row. Compensated
rounding, introduced by GPTQ~\cite{frantar2023gptq}, quantizes a working row
$V$ block by block from left to right. Once block $b$ is set to
$\widehat R_{4,b}$, the later columns $V_{>b}$ absorb its error by least
squares,
\begin{equation}
  V_{>b}\leftarrow V_{>b}-(V_b-\widehat R_{4,b})\,
    U_{H,bb}^{-1}U_{H,b,>b},
  \label{eq:construction-feedback}
\end{equation}
where $U_H$ is upper triangular with
$U_H^\top U_H=(H_{X_4}+\lambda_{\mathrm d}I)^{-1}$ and damping
$\lambda_{\mathrm d}>0$.

\textbf{Row targets.} $G$ couples the rows of $\Delta_R=R-R_4$. Fixing all
rows except $i$ and completing the square gives
\begin{equation}
  \begin{aligned}
    \mathcal J_{\mathrm{wt}}
      &=g_{ii}\|\tilde R_i-R_{4,i}\|_{H_{X_4}}^2+C_i,\\
    \tilde R_i
      &=R_i+\tfrac1{g_{ii}}\sum\nolimits_{j\ne i}g_{ij}\Delta_{R,j},
  \end{aligned}
  \label{eq:shifted-target}
\end{equation}
where $\Delta_{R,j}$ is row $j$ of $\Delta_R$, $g_{ij}$ are the entries of
$G$, $\|u\|_H^2=uHu^\top$, and $C_i$ does not depend on $R_{4,i}$
(\cref{app:coupled}). Each row is thus fitted to a target shifted by the
current errors of the other rows, so that it partly offsets them.

\textbf{Search.} In the first pass of \cref{alg:joint-calibration}, every
scale candidate is rounded by compensated rounding from $V=\tilde R_i$ and
scored under the damped $H_{X_4}$. Later passes score the same candidates by
the exact change of the row cost under the undamped $H_{X_4}$.
Rows of a group are searched in parallel without seeing each other's new
changes, so changes that each lower their own row cost can still raise
$\mathcal J_{\mathrm{wt}}$ together. We therefore accept them greedily by
the exact change of $\mathcal J_{\mathrm{wt}}$, so it never increases
(\cref{app:coupled,app:calibration}).

\subsection{Kernel optimization}
\label{sec:kernel}
\label{sec:kernel-optimization}

\begin{figure}[htbp]
  \centering
  \includegraphics[width=\linewidth,pagebox=cropbox,clip]{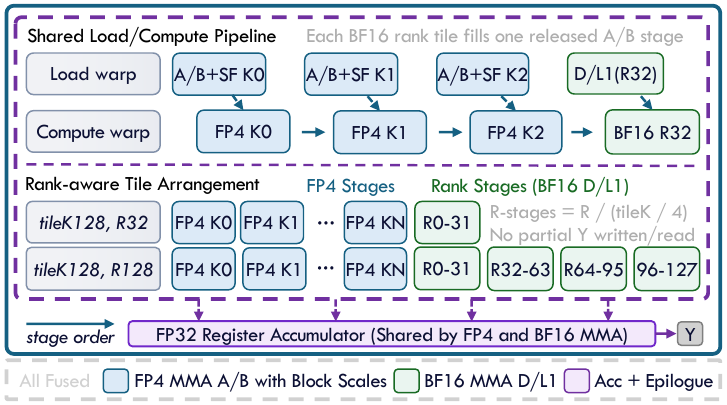}
  \caption{Fused NVFP4 residual and BF16 rank-up GEMM.}
  \label{fig:fused-kernel}
\end{figure}

For efficient deployment, we implement a fused kernel for the two-branch
layer.

\textbf{Shared FP32 accumulator.} The warp-specialized kernel fuses the
NVFP4 residual GEMM with the BF16 rank-up correction
(\cref{fig:fused-kernel}). Both compute phases
update one FP32 register accumulator, avoiding intermediate output traffic
to high-bandwidth memory. With both contributions in common accumulator
units, one FP32 scale--bias epilogue precedes BF16 conversion.

\textbf{Tile partitioning and layout.} A dedicated data-loading warp streams FP4 operands and block scales through a multistage
shared-memory ring. After the residual reduction loop, released stages hold
BF16 rank-reduced activations $D$ and output factor $L_1$ (figure notation).
Each BF16 tile fills one stage with $\mathrm{tile}_K/4$ rank columns,
appending $r/(\mathrm{tile}_K/4)$ stages for ranks divisible by
$\mathrm{tile}_K/4$. Autotuning jointly selects tile shape,
$\mathrm{tile}_K$, and operand orientation. Swaps for narrow output tiles
apply to both compute phases, preserving their shared accumulation layout.

\section{Experiments}
\label{sec:experiments}

\begin{table*}[t]
  \captionsetup{skip=3pt}
  \caption{Video generation results on HunyuanVideo-13B and Wan2.2-A14B.}
  \label{tab:evalcrafter-results}
  \centering
  \footnotesize
  \renewcommand{\arraystretch}{0.88}
  \setlength{\tabcolsep}{3pt}
  \begin{tabular*}{\textwidth}{@{\extracolsep{\fill}}l*{9}{r}@{}}
    \toprule
    \multicolumn{7}{@{}l}{\makebox[0pt][l]{\makebox[\textwidth][c]{\textbf{HunyuanVideo-13B/Wan2.2-A14B}}}} & \multicolumn{3}{c}{VBench} \\
    \cmidrule(l){8-10}
    Method & CLIPSIM$\uparrow$ & VQA(Avg)$\uparrow$ & $\Delta$FSCore$\downarrow$ & PSNR$\uparrow$ & SSIM$\uparrow$ & LPIPS$\downarrow$ & S.C.$\uparrow$ & A.Q.$\uparrow$ & I.Q.$\uparrow$ \\
    \midrule
    BF16 & 0.194 / 0.203 & 81.23 / 87.39 & -- & -- & -- & -- & 98.15 / 97.37 & 69.01 / 68.48 & 72.42 / 75.62 \\
    ViDiT-Q & 0.157 / 0.153 & 72.79 / 76.62 & 1.326 / 1.486 & 14.74 / 14.79 & 0.467 / 0.415 & 0.535 / 0.572 & 93.67 / 91.57 & 62.70 / 63.45 & 64.15 / 68.94 \\
    S2Q-VDiT & 0.174 / 0.165 & 74.97 / 83.33 & 0.795 / 0.976 & 18.77 / \textbf{18.26} & 0.613 / \textbf{0.552} & 0.306 / 0.278 & 95.80 / 94.66 & 65.10 / 67.13 & 67.83 / 73.81 \\
    DeltaQuant & 0.183 / 0.179 & 78.46 / 83.63 & 0.656 / 0.438 & 18.87 / 17.45 & 0.605 / 0.482 & 0.367 / 0.347 & 96.06 / 96.35 & 66.89 / 66.69 & 69.38 / 73.52 \\
    SVDQ & 0.163 / 0.150 & 73.61 / 77.30 & 1.453 / 1.317 & 15.29 / 15.63 & 0.496 / 0.436 & 0.502 / 0.549 & 91.78 / 93.28 & 62.95 / 64.37 & 65.67 / 70.09 \\
    H-SVDQ & 0.181 / 0.172 & 76.56 / 82.10 & 0.831 / 0.762 & 19.13 / 17.62 & 0.643 / 0.507 & 0.325 / 0.351 & 95.49 / 95.21 & 66.10 / 65.79 & 68.36 / 72.13 \\
    GPTAQ & 0.182 / 0.179 & 78.07 / 84.84 & 1.003 / 0.717 & 18.75 / 17.10 & 0.620 / 0.493 & 0.371 / 0.416 & 94.17 / 94.98 & 66.66 / 66.34 & 69.02 / 72.93 \\
    Ours & \textbf{0.192} / \textbf{0.196} & \textbf{79.89} / \textbf{86.79} & \textbf{0.437} / \textbf{0.244} & \textbf{20.88} / 18.13 & \textbf{0.678} / 0.531 & \textbf{0.273} / \textbf{0.265} & \textbf{97.43} / \textbf{96.86} & \textbf{68.35} / \textbf{67.90} & \textbf{71.54} / \textbf{74.88} \\
    \bottomrule
  \end{tabular*}
\end{table*}

\begin{table*}[t]
  \captionsetup{skip=3pt}
  \caption{Video generation results on MiniMax-H3 and MiniMax-H3 Fast.}
  \label{tab:h3-results}
  \centering
  \footnotesize
  \renewcommand{\arraystretch}{0.9}
  \setlength{\tabcolsep}{1.5pt}
  \begin{minipage}[t]{0.49\textwidth}
    \begin{tabular*}{\linewidth}{@{\extracolsep{\fill}}lrrrrrr@{}}
      \toprule
      \multicolumn{7}{c}{\textbf{MiniMax-H3 base, 50 steps}} \\
      \midrule
      Method & PSNR$\uparrow$ & SSIM$\uparrow$ & LPIPS$\downarrow$ & CLIP$\uparrow$ & VQA$\uparrow$ & Flow$\downarrow$ \\
      \midrule
      BF16 & -- & -- & -- & 0.2004 & 68.67 & -- \\
      RTN & 12.36 & 0.5411 & 0.485 & 0.2011 & 61.74 & 0.926 \\
      ViDiT-Q & 12.94 & 0.5522 & 0.487 & 0.2003 & 62.52 & 1.276 \\
      S2Q-VDiT & 9.55 & 0.3684 & 0.692 & 0.1901 & 39.15 & 3.364 \\
      DeltaQuant & 14.21 & 0.5759 & 0.419 & 0.2010 & 65.77 & 0.577 \\
      SVDQ & 14.04 & 0.5734 & 0.418 & 0.2023 & 65.87 & 0.629 \\
      H-SVDQ & 14.26 & 0.5865 & 0.410 & 0.1994 & 65.26 & 0.737 \\
      GPTAQ & 14.03 & 0.5799 & 0.413 & 0.2014 & 66.87 & 0.767 \\
      Ours & \textbf{14.55} & \textbf{0.5871} & \textbf{0.396} & \textbf{0.2029} & \textbf{67.33} & \textbf{0.397} \\
      \bottomrule
    \end{tabular*}
  \end{minipage}\hfill
  \begin{minipage}[t]{0.49\textwidth}
    \begin{tabular*}{\linewidth}{@{\extracolsep{\fill}}lrrrrrr@{}}
      \toprule
      \multicolumn{7}{c}{\textbf{MiniMax-H3 Fast, 4 steps}} \\
      \midrule
      Method & PSNR$\uparrow$ & SSIM$\uparrow$ & LPIPS$\downarrow$ & CLIP$\uparrow$ & VQA$\uparrow$ & Flow$\downarrow$ \\
      \midrule
      BF16 & -- & -- & -- & 0.1922 & 64.15 & -- \\
      RTN & 12.60 & 0.4107 & 0.610 & 0.1923 & 61.94 & 0.749 \\
      ViDiT-Q & 13.24 & 0.4299 & 0.591 & \textbf{0.1934} & 61.93 & 0.661 \\
      S2Q-VDiT & 12.60 & 0.4306 & 0.649 & 0.1871 & 48.02 & 2.728 \\
      DeltaQuant & 13.95 & 0.4562 & 0.535 & 0.1897 & 63.12 & 0.456 \\
      SVDQ & 13.87 & 0.4517 & 0.539 & 0.1909 & 63.27 & \textbf{0.401} \\
      H-SVDQ & 13.75 & 0.4464 & 0.546 & 0.1915 & 63.71 & 0.546 \\
      GPTAQ & 13.98 & 0.4559 & 0.546 & 0.1917 & 63.01 & 0.474 \\
      Ours & \textbf{14.12} & \textbf{0.4640} & \textbf{0.524} & 0.1922 & \textbf{64.50} & 0.502 \\
      \bottomrule
    \end{tabular*}
  \end{minipage}
\end{table*}

\begin{table*}[t]
  \captionsetup{skip=3pt}
  \caption{Image generation results on Qwen-Image with 4 steps.}
  \label{tab:qwen-image-results}
  \label{tab:qwen-text-results}
  \centering
  \footnotesize
  \renewcommand{\arraystretch}{0.9}
  \setlength{\tabcolsep}{4pt}
  \begin{tabular*}{\textwidth}{@{\extracolsep{\fill}}l@{\hspace{8pt}}rrrr@{\hspace{10pt}}rrrr@{\hspace{10pt}}rrr@{}}
    \toprule
    & \multicolumn{4}{c}{MJHQ} & \multicolumn{4}{c}{sDCI} & \multicolumn{3}{c}{LongText-Bench} \\
    \cmidrule(lr){2-5}\cmidrule(lr){6-9}\cmidrule(l){10-12}
    Method & PSNR$\uparrow$ & LPIPS$\downarrow$ & CLIP-T$\uparrow$ & Reward$\uparrow$ & PSNR$\uparrow$ & LPIPS$\downarrow$ & CLIP-T$\uparrow$ & Reward$\uparrow$ & PSNR$\uparrow$ & LPIPS$\downarrow$ & Acc.$\uparrow$ \\
    \midrule
    BF16 & -- & -- & 29.421 & 1.360 & -- & -- & 26.463 & 1.292 & -- & -- & 92.212 \\
    RTN & 15.50 & 0.348 & 29.042 & 1.284 & 14.39 & 0.361 & 26.216 & 1.261 & 12.84 & 0.364 & 90.635 \\
    ViDiT-Q & 15.08 & 0.308 & 28.627 & 1.232 & 14.08 & 0.358 & 26.136 & 1.259 & 10.07 & 0.531 & 83.812 \\
    S2Q-VDiT & 12.29 & 0.540 & 28.565 & 1.249 & 11.95 & 0.561 & 25.783 & 1.249 & 10.49 & 0.575 & 84.324 \\
    SVDQ & 15.62 & 0.296 & 29.346 & 1.281 & 14.52 & 0.345 & 26.296 & 1.262 & 12.87 & 0.341 & 90.599 \\
    H-SVDQ & 15.83 & 0.290 & 29.266 & 1.306 & 14.53 & 0.330 & 26.451 & 1.316 & 12.70 & 0.353 & 91.373 \\
    GPTAQ & 15.73 & 0.293 & 29.143 & 1.259 & 14.60 & 0.328 & 26.319 & 1.299 & 12.71 & 0.352 & 89.637 \\
    Ours & \textbf{16.47} & \textbf{0.282} & \textbf{29.811} & \textbf{1.341} & \textbf{15.05} & \textbf{0.308} & \textbf{26.603} & \textbf{1.339} & \textbf{12.95} & \textbf{0.334} & \textbf{91.837} \\
    \bottomrule
  \end{tabular*}
\end{table*}

\subsection{Experimental setup}
\label{sec:evaluation-scope}

\textbf{Models and baselines.}
We evaluate the video models HunyuanVideo-13B, Wan2.1-14B, Wan2.2-A14B, MiniMax-H3 with 50 steps and MiniMax-H3 Fast with 4 steps~\cite{minimax2026h3,fastvideo2026fasth3dense}, and the image model Qwen-Image~\cite{qwen2025image} with 50 steps or 4 steps (Qwen-Image-Lightning V2.0 adapter~\cite{lightx2v2025lightning}). The baselines are RTN, ViDiT-Q~\cite{zhao2025viditq}, S2Q-VDiT~\cite{feng2025s2qvdit}, DeltaQuant~\cite{li2026deltaquant}, SVDQ~\cite{li2025svdquant,frantar2023gptq}, H-SVDQ~\cite{wang2026hsvdquant}, and GPTAQ~\cite{li2025gptaq}. SVDQ and GPTAQ allocate the bypass by weight SVD and then calibrate the residual with GPTQ and GPTAQ, respectively. All methods use the same NVFP4 weight and activation format, and low-rank methods use rank-64 bypasses. These are our adaptations to a common format, not the published configurations.

\textbf{Evaluation.}
Videos are scored with EvalCrafter and VBench~\cite{liu2024evalcrafter,huang2024vbench}, and images on MJHQ, sDCI, and English LongText-Bench~\cite{playground2024mjhq,urbanek2024dci,xomni2025longtext}. Each quantized output is paired with the BF16 output from the same prompt and initial noise, which is the reference for PSNR, SSIM, and LPIPS. In the tables, S.C., A.Q., and I.Q.\ are the VBench subject consistency, aesthetic quality, and imaging quality, CLIP, VQA, and Flow abbreviate CLIPSIM, VQA(Avg), and $\Delta$FSCore, Reward and Acc.\ are ImageReward and TextAcc, and bold marks the best quantized result. Across all experiments, we use 25\% of the data for calibration and the remaining 75\% for evaluation. Protocols and metric definitions are in \cref{app:video-evaluation,app:image-evaluation}.

\subsection{Results}
\label{sec:main-results}
\label{sec:video-results}
\label{sec:h3-results}
\begin{figure}[t]
  \centering
  \captionsetup{skip=3pt}
  \captionsetup[subfigure]{justification=centering,skip=2pt}
  \begin{subfigure}[t]{0.49\linewidth}
    \centering
    \includegraphics[width=\linewidth]{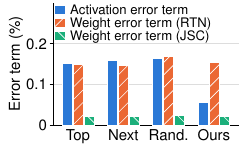}
    \caption{Bypass error terms}
    \label{fig:bypass-choice}
  \end{subfigure}\hfill
  \begin{subfigure}[t]{0.49\linewidth}
    \centering
    \includegraphics[width=\linewidth]{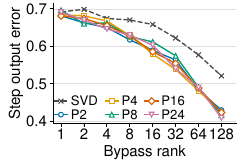}
    \caption{Rank and probes}
    \label{fig:rank-probe-allocation}
  \end{subfigure}
  \caption{Error terms, bypass rank, and probe count.}
  \label{fig:analysis-comparison}
\end{figure}

\newsavebox{\fasthGrid}
\newcommand{\fasthScale}{0.8372807}
\begin{figure*}[tp]
  \centering
  \sbox{\fasthGrid}{\input{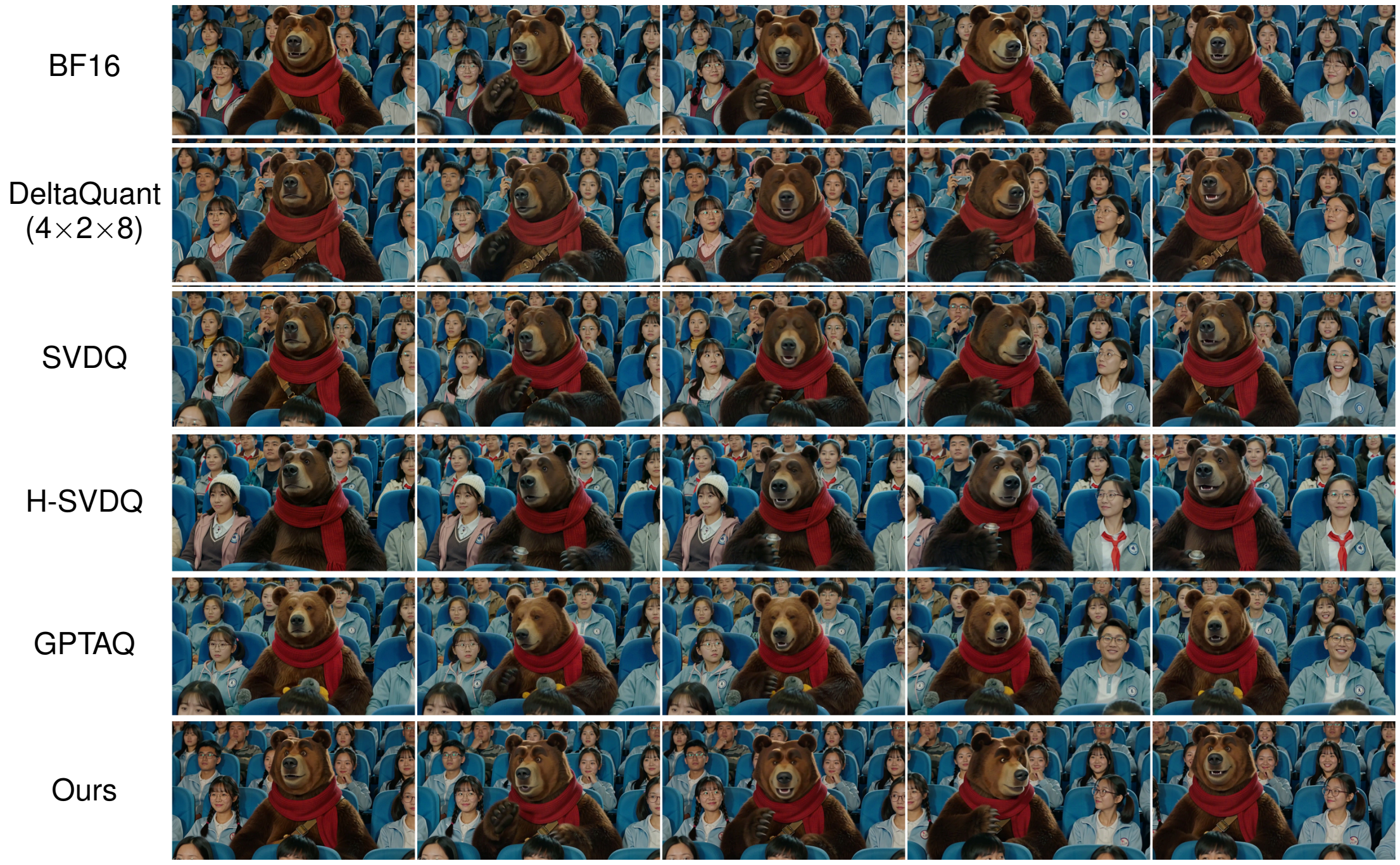}}%
  \newcommand{\fasthRow}[2]{%
    \clipbox{{0.0059289\width} {#1\height} {0.0027668\width} {#2\height}}{\usebox{\fasthGrid}}}%
  \vbox{\offinterlineskip
    \fasthRow{0.8886972}{0.0030198}\vskip 1.5pt
    \fasthRow{0.4456428}{0.4456428}\vskip 1.5pt
    \fasthRow{0.3352028}{0.5565142}\vskip 1.5pt
    \fasthRow{0.2243313}{0.6669543}\vskip 1.5pt
    \fasthRow{0.1134599}{0.7778257}\vskip 1.5pt
    \fasthRow{0.0030198}{0.8886972}}
  \caption{MiniMax-H3 Fast: a bear wearing a red scarf sits beside a person in a theater. Columns show frames 1, 32, 63, 93, and 124.}
  \label{fig:h3-dist-qualitative}
\end{figure*}

\begin{figure*}[tp]
  \centering
  \begin{minipage}{\textwidth}
  \centering
  \input{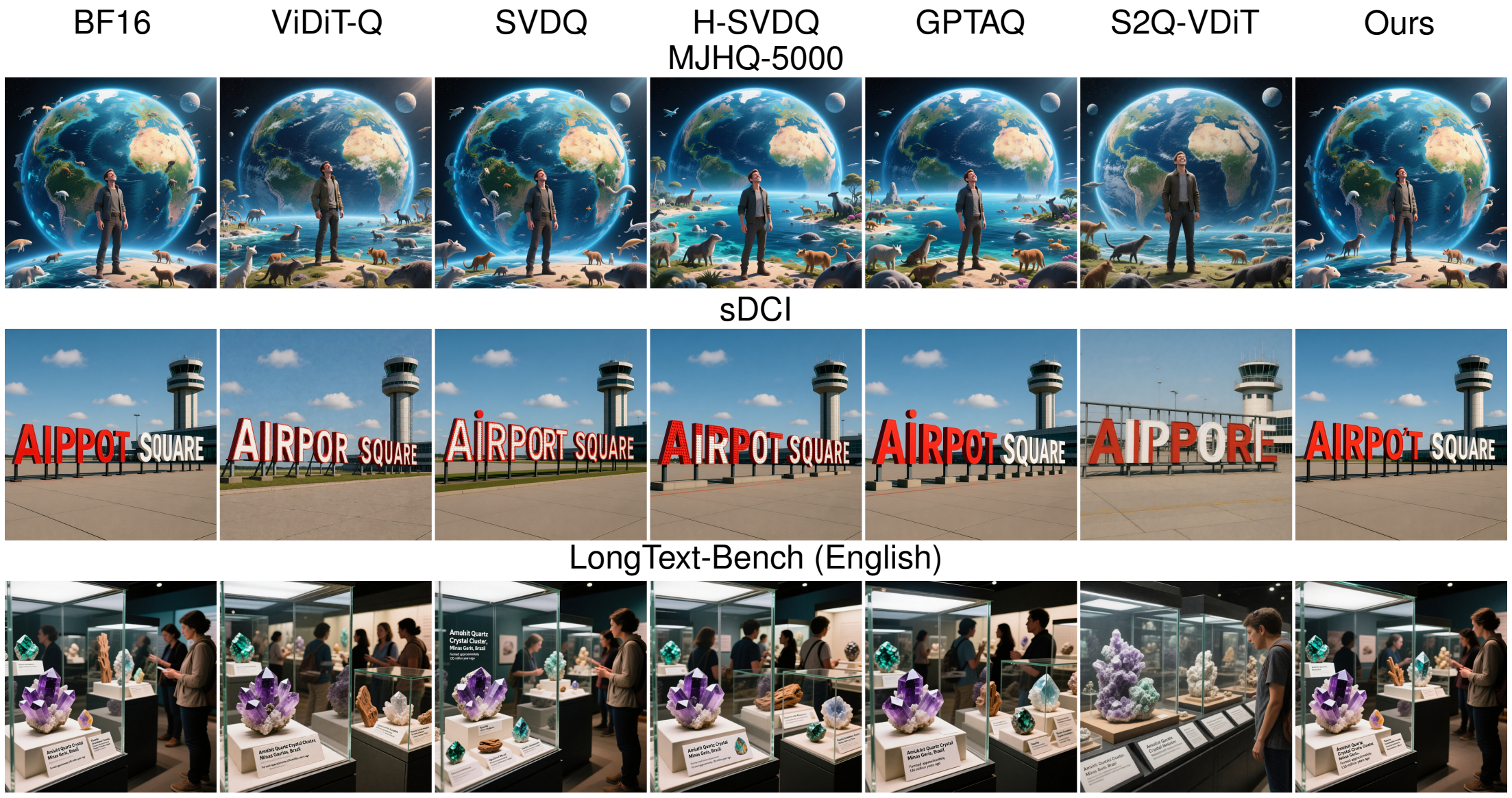}
  \caption{Qwen-Image with 4 steps.}
  \label{fig:qwen-dist-main-qualitative}
  \end{minipage}
\end{figure*}

\textbf{Video generation.}
On HunyuanVideo and Wan2.2 (\cref{tab:evalcrafter-results}), KLI-Quant is best among quantized methods in 10 of the 12 quality and fidelity comparisons and in all six VBench comparisons. It lowers LPIPS by 4.7--10.8\% relative to the strongest baseline on each model. The exceptions are PSNR and SSIM on Wan2.2, where S2Q-VDiT is higher. Additional Wan2.1-14B results appear in \cref{tab:wan21-results}. On MiniMax-H3 (\cref{tab:h3-results}), KLI-Quant is best in all six metrics with 50 steps and in four with 4 steps. On MiniMax-H3 Fast, ViDiT-Q has the highest CLIPSIM and SVDQ the lowest flow-score deviation. \Cref{fig:h3-dist-main-qualitative,fig:h3-dist-qualitative} and \cref{app:qualitative-video} show paired examples.

\textbf{Image generation.}
\label{sec:qwen-results}
On Qwen-Image with 4 steps (\cref{tab:qwen-image-results}), KLI-Quant is best in all 11 quantized comparisons. Its LongText-Bench TextAcc is 91.837\%, 0.375 percentage points below BF16. With 50 steps (\cref{tab:qwen-image-base50}), KLI-Quant is best in 9 of the 11, including PSNR and LPIPS on every dataset, and SVDQ leads in sDCI CLIP-T and ImageReward. \Cref{fig:qwen-dist-main-qualitative} shows one 4-step example per dataset, and \cref{app:qualitative-video} gives more.

\textbf{Language models.}
\label{sec:llm-results}
On Qwen3-8B (\cref{app:llm-results}), KLI-Quant leads quantized methods on 8 of 9 metrics, reaching 72.53\% MMLU-Pro accuracy (SVDQ: 64.16\%; BF16: 74.94\%) and 9.732 WikiText-2 perplexity (BF16: 9.730).

\textbf{Kernel performance.}
\label{sec:kernel-results}
Across 14 GEMM shapes, the fused kernel with the rank-64 BF16 bypass has a
latency of $1.00$--$1.15\times$ that of a direct W4A4 GEMM without the
bypass in tuned CUTLASS ($1.04$--$1.24\times$ against cuBLASLt), and it is
$3.12\times$ faster than BF16 on average. GEMM and end-to-end latencies
appear in \cref{app:latency-details}.

\subsection{Ablation studies and analysis}
\label{sec:ablation}
\label{sec:parameter-study}
\label{sec:allocation-efficiency}

\begin{table}[t]
  \captionsetup{skip=3pt}
  \caption{Independent component ablations on MiniMax-H3 Fast.}
  \label{tab:ablation}
  \centering
  \footnotesize
  \renewcommand{\arraystretch}{0.9}
  \setlength{\tabcolsep}{3pt}
  \begin{tabular*}{\linewidth}{@{\extracolsep{\fill}}llcccc@{}}
    \toprule
    Bypass $L$ & Sensitivity & Residual & \shortstack{Step Output\\Error$\downarrow$} & SSIM$\uparrow$ & LPIPS$\downarrow$ \\
    \midrule
    Act.\ term & $G$ & JSC & 0.09736 & 0.4640 & 0.524 \\
    Act.\ term & $G$ & GPTQ & 0.1046 & 0.4601 & 0.522 \\
    Act.\ term & $G=I$ & JSC & 0.1159 & 0.4583 & 0.531 \\
    Weight SVD & $G$ & JSC & 0.1223 & 0.4517 & 0.539 \\
    \bottomrule
  \end{tabular*}
\end{table}

\textbf{Ablation.}
\Cref{tab:ablation} independently removes one component from the full method on MiniMax-H3 Fast (rank 64; four-step mean step output error). Replacing JSC with GPTQ retains the $G$-weighted activation-term allocation; removing sensitivity sets $G=I$ in both allocation and JSC; replacing activation-term allocation with weight SVD retains $G$-weighted JSC. Relative to these variants, the full method lowers step output error by 6.9\%, 16.0\%, and 20.4\%, respectively. It achieves the highest SSIM (0.4640); the GPTQ variant has slightly lower LPIPS (0.522 vs.\ 0.524). The separate rank-128 comparison of complete methods also gives KLI-Quant the lowest step output error at every step (\cref{fig:posthead-mse}).

\textbf{Activation and weight error reduction.}
\label{sec:bypass-choice}
\Cref{fig:bypass-choice} splits the layer output error into the activation term and the weight term of \cref{eq:three-term-identity} on 36 layers, for four rank-64 bypasses: weight-SVD directions 1--64 (the SVDQ choice), 65--128, 64 randomly selected weight-SVD directions, and the Stage~1 bypass. Each bar is the median over layers of a $G$-weighted term divided by $\|G^{1/2}WH_{X_4}^{1/2}\|_F^2$. With the SVDQ bypass and RTN residuals, the two terms are about equal (0.151\% and 0.150\%), as in \cref{sec:motivation-activation-weight}. Each stage of KLI-Quant lowers the term it targets. Stage~1 lowers the activation term in all 36 layers, to 0.056\%, at least $2.7\times$ below the weight-SVD bypasses. JSC lowers the weight term about $7\times$ from RTN, to 0.021--0.023\%. Across the alternative bypasses, the achieved weight cost varies a median 0.05 times as much as the activation cost, supporting \cref{asm:floor} for these candidates. JSC uses $H_{X_4}$ and the full downstream metric $G$.

\textbf{Bypass rank and probe count.}
\Cref{fig:rank-probe-allocation} varies the rank from 1 to 128 and the number of probes for $G$ from 2 to 24 (P2--P24) at step 2. Only the residual input is 4-bit, so the layer output error is the activation term $\Delta_X(W-L)^\top$. Task-sensitive allocation has a lower step output error than unweighted SVD in 39 of 40 matched-rank comparisons (all except rank 1, P8). At rank 64, every probe setting is 5.89--7.59\% below SVD at rank 128. The probe count matters less than the rank: the five settings differ by 1.8\% at rank 64 and 5.2\% at rank 128, with no consistent gain from more probes. We use rank 64 and two probes (details in \cref{app:rank-probe-allocation}).

\section{Conclusion}
\label{sec:conclusion}

KLI-Quant performs W4A4 post-training quantization for diffusion models
by weighting layer output error according to its effect on step output
error. Its decomposition separates activation and weight terms: bypass
allocation targets activation error, while JSC calibrates residual block
scales and 4-bit codes under quantized inputs. Against seven W4A4
baselines in the same NVFP4 format, KLI-Quant leads most quality and fidelity comparisons across the evaluated models. Over BF16, the fused kernels give $3.12\times$ average GEMM and
$1.93\times$ end-to-end speedups on MiniMax-H3.

\clearpage
\makeatletter\global\@colroom\@colht\global\vsize\@colroom\makeatother
\floatsep=\kliqfs \textfloatsep=\kliqtfs
\dblfloatsep=\kliqdfs \dbltextfloatsep=\kliqdtfs
\setcounter{dbltopnumber}{2}

\section*{LLM Usage Statement}
\label{sec:llm-usage}
OpenAI Codex was used to assist with manuscript writing and editing, literature and citation checks, figure-generation scripts, LaTeX preparation, and the implementation and debugging of experimental code. The tool provided editable text and code drafts and suggestions across these activities. The authors made the final research decisions and take full responsibility for the paper's content, methods, experimental results, citations, and claims.

\AddToHook{file/4_appendix.tex/before}{\raggedbottom}
\AddToHook{file/4_appendix.tex/after}{\flushbottom}
\AddToHook{env/equation/begin}{%
  \setlength{\abovedisplayskip}{6pt minus 1pt}%
  \setlength{\belowdisplayskip}{6pt minus 1pt}%
  \setlength{\abovedisplayshortskip}{3pt minus 1pt}%
  \setlength{\belowdisplayshortskip}{4pt minus 1pt}%
}
\clearpage
\appendix
\numberwithin{equation}{section}
\maketitlesupplementary
\section{Appendix: derivations and implementation details}
\label{sec:method-appendix}

\newcounter{kliqappendixfigure}
\renewcommand{\thekliqappendixfigure}{A.\arabic{kliqappendixfigure}}
\crefalias{kliqappendixfigure}{figure}

This appendix gives the derivations for \cref{sec:error-modeling}, the
analysis of the two stages in \cref{sec:allocation-calibration}, and the
formulas and numerical details of the Stage~2 procedure
(\cref{alg:joint-calibration}) to our Kit for Lean Inference quantization method.

\subsection{Downstream sensitivity}
\label{app:sensitivity}

This section details the three approximations behind
\cref{eq:step-error-chain}. We treat $f$ as a flattened vector, write
$y_t\in\mathbb R^o$ for the layer output of token $t$, and set
$J_t=\partial f/\partial y_t$.
We collect each layer's input $X$ from the frozen reference model, so
the calibration objective omits changes to $X$ caused by quantizing
preceding layers.

\emph{First-order linearization.} Keeping the first-order term of a Taylor
expansion around the reference computation gives
\begin{equation}
  \delta f\approx\sum_tJ_te_t.
  \label{eq:first-order-response}
\end{equation}

\emph{Cross-token and cross-layer interactions.} Squaring the linearized
response gives
\begin{equation}
  \frac1m\Bigl\|\sum\nolimits_tJ_te_t\Bigr\|_2^2
   =\sum_te_t^\top G_te_t
    +\frac1m\sum_{t\ne t'}e_t^\top J_t^\top J_{t'}e_{t'}.
  \label{eq:cross-token-response}
\end{equation}
We keep the per-token terms and omit interactions between distinct tokens,
and likewise between layers. Token mixing downstream and shared weight
errors can make these interactions nonzero; omitting them makes a layer-wise
channel metric tractable.

\emph{Averaged sensitivity.} Replacing the token-dependent $G_t$ by the
common metric $\bar G$, the second step of \cref{eq:step-error-chain},
neglects the statistical dependence between $G_t$ and $e_te_t^\top$. This
separation of sensitivity from error statistics follows the factored
curvature model of convolutional K-FAC~\cite{grosse2016kfc}, where activation
and derivative statistics are factored and spatial derivative correlations
are neglected. \Cref{fig:mlp-sensitivity-ranking} tests whether the resulting
frozen layer metric predicts which equal-energy perturbations at a single
token most affect the denoiser output.

\emph{Normalization.} Both stages are invariant to a positive rescaling of
$G$, so the normalization by $\alpha=\operatorname{tr}(\bar G)/o$ only sets
the relative size of the regularizer $\rho$.

\subsection{Two exact error decompositions}
\label{app:decomposition}

Both decompositions start from the residual-branch error
$E=XR^\top-X_4R_4^\top$. Applied to the residual $R$, the proof of
Proposition~4.1 in SVDQuant~\cite{li2025svdquant} adds and subtracts
$XR_4^\top$, giving
\begin{equation}
  E=X(R-R_4)^\top+(X-X_4)R_4^\top=X\Delta_R^\top+\Delta_XR_4^\top.
  \label{eq:alternative-error-split}
\end{equation}
Adding and subtracting $X_4R^\top$ gives our decomposition,
\begin{equation}
  E=(X-X_4)R^\top+X_4(R-R_4)^\top=\Delta_XR^\top+X_4\Delta_R^\top.
  \label{eq:appendix-error-derivation}
\end{equation}
Both identities are exact. They assign the mixed product
$\Delta_X\Delta_R^\top$ to different terms.

With the bypass and calibration inputs fixed, updating $R_4$ also changes
the activation term $\Delta_XR_4^\top$ in \cref{eq:alternative-error-split}.
In \cref{eq:appendix-error-derivation}, Stage~1 fixes the activation term
$\Delta_XR^\top$ when it selects the bypass $L$ and hence $R=W-L$.
Stage~2 can then calibrate $R_4$ against the weight error $X_4\Delta_R^\top$
on the quantized inputs the residual branch receives. The interaction between
the two terms in the squared objective is examined in \cref{app:interactions}.

\subsection{The activation--weight interaction}
\label{app:interactions}
\label{app:interaction-bound}

Expanding the square in \cref{eq:downstream-surrogate} with
$E=E_{\mathrm{act}}+E_{\mathrm{wt}}$ gives the cross term
$\mathcal J_\times$ of \cref{eq:three-term-identity},
\begin{equation}
  \mathcal J_\times
   =\frac2N\bigl\langle E_{\mathrm{act}}G^{1/2},E_{\mathrm{wt}}G^{1/2}
     \bigr\rangle_F
   =2\operatorname{tr}\bigl(GRH_{\Delta_XX_4}\Delta_R^\top\bigr),
  \label{eq:cross-term}
\end{equation}
with $H_{\Delta_XX_4}=\Delta_X^\top X_4/N$. The first two terms are
nonnegative, and the signed $\mathcal J_\times$ measures how the two error
sources reinforce or cancel. The expansion requires no independence or zero-mean assumption about
rounding. We drop $\mathcal J_\times$ because it is small in our
measurements. Fitting it would also
reward a weight error that opposes the activation error on the calibration
inputs.

\textbf{Measured cross term.}
For each linear layer of MiniMax-H3 Fast (G64/P2), we average the signed
cross term $\mathcal J_\times$ and the total objective $\mathcal J$ in
\cref{eq:three-term-identity} separately over 52 prompts and all four
denoising steps. The ratio of these means ranges from $-1.25\%$ to
$+0.60\%$ across the 312 layers and is negative in 309 of them.
The aggregate cross term is $-0.78\%$ of the total objective.

Applying Cauchy--Schwarz to the inner product in \cref{eq:cross-term} gives
\begin{equation}
  \mathcal J \le
  \bigl(\sqrt{\mathcal J_{\mathrm{act}}}+\sqrt{\mathcal J_{\mathrm{wt}}}\bigr)^2.
  \label{eq:local-component-bound}
\end{equation}
For fixed component energies, the bound is attained when the two
$G$-weighted errors point in the same direction.

\subsection{Separating the two stages}
\label{app:floor}

Let $L_1$ minimize $\mathcal J_{\mathrm{act}}$ and let $L^\star$ minimize
$\mathcal J_{\mathrm{act}}(L)+\Phi_{\mathrm W}(L)$, the objective
\cref{eq:two-component-objective} after minimizing over $R_4$. Since
$\mathcal J_{\mathrm{act}}(L_1)\le\mathcal J_{\mathrm{act}}(L^\star)$,
\begin{equation}
  \begin{aligned}
    \mathcal J_{\mathrm{act}}(L_1)+\Phi_{\mathrm W}(L_1)
     &\le\mathcal J_{\mathrm{act}}(L^\star)+\Phi_{\mathrm W}(L^\star)\\
     &\quad+\bigl[\Phi_{\mathrm W}(L_1)-\Phi_{\mathrm W}(L^\star)\bigr].
  \end{aligned}
  \label{eq:floor-gap}
\end{equation}
With exact minimization in both stages, \cref{eq:floor-gap} bounds the
objective gap by the difference in attainable weight costs at $L_1$ and
$L^\star$.

To assess \cref{asm:floor}, we compared the Stage~1 bypass with each of
the three rank-64 weight-SVD bypasses in \cref{fig:bypass-choice}, using
the same JSC procedure for all four. In all 36 layers, the bypass selected
by $\mathcal J_{\mathrm{act}}$ still had the lowest measured
$\mathcal J_{\mathrm{act}}+\mathcal J_{\mathrm{wt}}$ among the four
candidates. The $0.05$ ratio in \cref{sec:ablation} is the median over layers of the
weight-cost range divided by the activation-cost range among the three
weight-SVD bypasses alone, whereas here we compare each of those
bypasses directly with the Stage~1 choice.
Across the 108 comparisons, the absolute change in
achieved weight cost was a median of $0.63\%$ of the activation-cost
reduction, with a maximum of $31.3\%$. In a separate MiniMax-H3 Fast
measurement (\cref{app:interactions}), the signed activation--weight
interaction was $-0.78\%$ of the total weighted objective in aggregate.
This does not establish that including the interaction would preserve the
bypass ordering in each layer.

\subsection{Optimality of the allocation}
\label{app:allocation}

This argument assumes exact leading singular vectors and exact arithmetic.
Symmetric positive-semidefinite square roots make each metric enter once
after squaring: if $H_{\Delta_X}=V\operatorname{diag}(\kappa)V^\top$ with
$\kappa\ge0$, then $H_{\Delta_X}^{1/2}=V\operatorname{diag}(\sqrt\kappa)
V^\top$. Hence, with $F(M)=G^{1/2}MH_{\Delta_X}^{1/2}$,
$\operatorname{tr}(GMH_{\Delta_X}M^\top)=\|F(M)\|_F^2$ for any $M$. With
$M=W-L$ this is $\mathcal J_{\mathrm{act}}(L)$, and linearity of $F$ gives
\cref{eq:allocation-objective}. Write $F_r=U_rU_r^\top F(W)$ for the target
in \cref{eq:allocation-target}. For any $L$ with
$\operatorname{rank}(L)\le r$, the matrix $F(L)$ has rank at most $r$, so
$\mathcal J_{\mathrm{act}}(L)$ is at least the best rank-$r$ approximation
error $\|F(W)-F_r\|_F^2=\sum_{k>r}\sigma_k^2$ (Eckart--Young). The factors
in \cref{eq:bypass-factors} give $F(BA)=F_r$ and attain this bound. Because
$U_rU_r^\top$ is an orthogonal projection, $F_r$ and $F(W)-F_r$ have zero
Frobenius inner product, so the loss also equals
$\|F(W)\|_F^2-\|U_r^\top F(W)\|_F^2$, which needs only $U_r$.

If $H_{\Delta_X}$ is invertible, $F$ is a rank-preserving bijection, so for
$\sigma_r>\sigma_{r+1}$ the minimizer is unique and equals
$G^{-1/2}F_rH_{\Delta_X}^{-1/2}=BA$. If $H_{\Delta_X}$ is singular, the
condition $F(L)=F_r$ fixes $L$ only on the range of $H_{\Delta_X}$. The
minimizer is then not unique, and $BA$ is still one. Computing $BA$ inverts
only $G\succ0$, so a singular Hessian causes no obstruction. Positive
rescaling of either
metric leaves $U_r$ and $BA$ unchanged, so $H_{\Delta_X}$ may be divided by
its mean diagonal during the solve.

\subsection{Stage 2 procedure and row coupling}
\label{app:coupled}

Let $\Delta_i=R_i-R_{4,i}$ denote row $i$ of $\Delta_R$, and set
$\Psi=G\Delta_R$. With $G\succ0$ and $H=H_{X_4}\succeq0$, the weight
cost is
\begin{equation}
  \mathcal J_{\mathrm{wt}}=\sum_{i,j}g_{ij}\Delta_iH\Delta_j^\top.
  \label{eq:full-weight-cost}
\end{equation}

\emph{Shifted target.} Holding the other rows fixed, the terms involving
row $i$ are $g_{ii}\Delta_iH\Delta_i^\top+2\Delta_iHc_i^\top$, where
$c_i=\sum_{j\ne i}g_{ij}\Delta_j$. Since $g_{ii}>0$, completing the
square gives
\begin{equation}
  g_{ii}\Delta_iH\Delta_i^\top+2\Delta_iHc_i^\top
   =g_{ii}\Bigl\|\Delta_i+\frac{c_i}{g_{ii}}\Bigr\|_H^2
    -\frac{c_iHc_i^\top}{g_{ii}},
  \label{eq:complete-square}
\end{equation}
The last term is constant in $R_{4,i}$, so the shifted target in
\cref{eq:shifted-target} is $\tilde R_i=R_i+c_i/g_{ii}$. Since
$\Psi_i=g_{ii}\Delta_i+c_i$, we compute it as
$\tilde R_i=R_{4,i}+\Psi_i/g_{ii}$.

\emph{Row groups.} Each pass visits the row groups in sequence. At the
start of a group $\mathcal I$, we form all its shifted targets from the
current $\Psi$ and search its rows in parallel (\cref{app:calibration}).

\emph{Exact group acceptance (line~\ref{alg:jsc-accept}).}
Let $\delta_i=R_{4,i}^{\mathrm{new}}-R_{4,i}$ be the change proposed
for row $i$ relative to the start of the group. Applying any subset
$\mathcal A\subseteq\mathcal I$ of these feasible updates replaces
$\Delta_i$ by $\Delta_i-\delta_i$ for $i\in\mathcal A$. Expanding
\cref{eq:full-weight-cost} about the start of the group gives
\begin{equation}
  \begin{aligned}
    \Delta\mathcal J_{\mathrm{wt}}(\mathcal A)
      &=\sum_{i\in\mathcal A}a_i
       +\sum_{\substack{i,j\in\mathcal A\\ i\ne j}}
          g_{ij}\,\delta_iH\delta_j^\top,\\
    a_i&=g_{ii}\,\delta_iH\delta_i^\top-2\,\delta_iH\Psi_i^\top.
  \end{aligned}
  \label{eq:group-acceptance}
\end{equation}
Here $a_i$ is the change in $\mathcal J_{\mathrm{wt}}$ for row $i$
alone. Starting with $\mathcal K=\varnothing$, we consider rows in
nondecreasing order of $a_i$ and add $i$ to $\mathcal K$ if its
marginal change is negative,
\begin{equation}
  a_i+2\sum_{j\in\mathcal K}g_{ij}\,\delta_iH\delta_j^\top<0.
  \label{eq:greedy-increment}
\end{equation}
Summing the accepted increments shows that a nonempty $\mathcal K$
strictly lowers $\mathcal J_{\mathrm{wt}}$. An empty set leaves the cost
unchanged. The update $\Psi\leftarrow\Psi-G_{:,\mathcal K}\delta_{\mathcal K}$
preserves $\Psi=G\Delta_R$ for the next group.

\emph{Termination (line~\ref{alg:jsc-stop}).} Calibration stops if a pass
accepts no row or the relative decrease in $\mathcal J_{\mathrm{wt}}$
over the pass falls below $\tau$.

\emph{Cost.} A pass visits $\lceil o/n_{\mathrm g}\rceil$ groups.
Scoring their proposals uses the products $\delta_iH$ and the pairwise
terms in \cref{eq:group-acceptance}. The dense updates of $\Psi$ cost
$O(o^2d)$ per pass in the worst case.

\subsection{Calibration details}
\label{app:calibration}
\label{app:representation}

\emph{Initialization.} To estimate $H=H_{X_4}$, we quantize each complete
calibration input tensor before sampling tokens. We initialize
the calibration proxy $P$ to the residual $R$ and obtain $R_4$ by absmax
scaling and nearest-code rounding (\cref{eq:motivation-nvfp4}). The tensor
scale $T$ stays fixed.

\emph{Block scoring (lines~\ref{alg:jsc-round}--\ref{alg:jsc-keep}).}
Let $d_0$ be the reconstruction of block $b$ in row $i$ on entry to the
block. For a candidate reconstruction $d$ with $\Gamma=d-d_0$, the change
in the block score is
\begin{equation}
  \Delta f=f_{i,b}(d)-f_{i,b}(d_0)
    =-2\langle\Gamma,h_0\rangle+\Gamma M\Gamma^\top,
  \label{eq:block-proposal-score}
\end{equation}
where $M$ and $h_0$ depend on the pass. For each candidate scale, we
score the initial reconstruction, then refine its codes in forward and
reverse order. The sweep starts at that reconstruction with
$h=h_0-\Gamma M$. Changing reconstructed coordinate $j$ by $\delta_j$
gives the score change and, upon acceptance, the update
\begin{equation}
  \Delta f_j=-2\delta_jh_j+\delta_j^2M_{jj},\qquad
  h\leftarrow h-\delta_jM_{j,:}.
  \label{eq:coordinate-update}
\end{equation}
Each coordinate takes the feasible code with the largest decrease if one
exists. The lowest-scoring state among the incoming block and all initial
and refined candidates becomes the tentative block reconstruction.

\emph{Construction ($k=0$).} We initialize the working target
$V\leftarrow\tilde R_{\mathcal I}$ for each row group $\mathcal I$.
At block entry, the current target $V_{i,b}$ and the layer's damped
factor $U_H$ give
\begin{equation}
  M=(U_{H,bb}^\top U_{H,bb})^{-1},\qquad h_0=(V_{i,b}-d_0)M.
  \label{eq:construction-context}
\end{equation}
For each scale, we round $V_{i,b}$ in order, compensating the remaining
coordinates within the block. After selecting the block reconstruction,
we propagate its error to the remaining columns of $V$ using
\cref{eq:construction-feedback}.

\emph{Refinement ($k\ge1$).} Each candidate scale requantizes the same
incoming block proxy $P_{i,b}$. The shifted target $\tilde R_i$ is fixed
at group entry. The tentative row $\widehat R_{4,i}$ includes earlier
block updates in input order, with $d_0=\widehat R_{4,i,b}$ at entry to
the current block. Using the undamped $H$, we set
\begin{equation}
  M=H_{bb},\qquad h_0=(\tilde R_i-\widehat R_{4,i})H_{:,b}.
  \label{eq:refinement-context}
\end{equation}
With the other rows fixed, changing this block changes
$\mathcal J_{\mathrm{wt}}$ by $g_{ii}\Delta f$.
In both phases, the completed row proposals are accepted or rejected
by the group rule in \cref{app:coupled}.

\subsection{Observation details}
\label{app:observations}

\textbf{Observation 1.} We test six linear projections in each of blocks
30, 35, and 40. For each layer, prompt, and step, we group the eigenvectors
of $Y^\top Y$ for the BF16 layer output $Y$ into 16 equal-sized sets in
descending eigenvalue order. We map the same Gaussian coefficients into
each subspace in turn, scaling the perturbation to match the layer's direct
NVFP4 rounding error energy for that prompt and step.

For each subspace, we take the geometric mean of the step output MSE over
four prompts and four steps. Each layer curve in \cref{fig:direction} shows
these 16 values relative to their geometric mean, with angular position
indicating within-layer response rank. Permuting subspace labels within
each prompt and step gives $p<0.01$ in 17 of 18 layers. To assess
repeatability, we rank the 16 eigenvalue groups by response in each
half of a prompt split. The per-layer Spearman correlation between halves,
averaged over two splits, has a median of $0.80$ across layers.

\emph{Predicting downstream impact.} On held-out FastH3 prompts, we compare
the ranking predicted by frozen $G_\ell$ with the video post-head MSE
measured against the unperturbed teacher at the same state. At one uniformly
sampled token per case, we test eight left singular vectors of
$WH_{\Delta_X}^{1/2}$ spaced evenly in rank across its stable spectrum.
Neither the token nor the directions are chosen using $G_\ell$ or the
measured error. Each realized BF16 output change $d$ matches a common
squared norm within 0.5\%. The teacher ranks by $d^\top G_\ell d/\|d\|_2^2$,
while local order ranks the same directions by singular value.

\par\smallskip
\noindent\begin{minipage}{\linewidth}
  \refstepcounter{kliqappendixfigure}\label{fig:mlp-sensitivity-ranking}%
  \includegraphics[width=\linewidth]{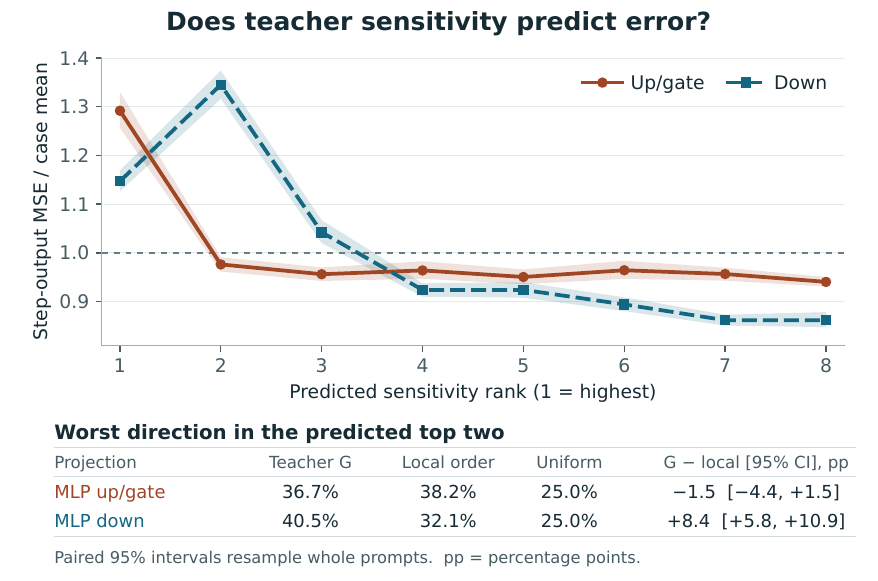}\par
  \smallskip
  {\small\noindent Figure~\thekliqappendixfigure. Step output MSE by predicted sensitivity rank.
\par}
\end{minipage}
\par\smallskip

At each predicted rank, we divide the MSE of that direction by the mean
over all eight directions in the same case, then average over steps,
blocks, and prompts. The top-two hit rate is the fraction of eligible
cases in which either of the two highest-ranked directions yields the
largest measured MSE. We resample whole prompts 10,000 times to obtain
95\% percentile intervals: pointwise for the curves and paired for
differences in hit rates.

For up/gate, the first choice of $G_\ell$ produces $1.292\times$ the
case-average MSE, but its top two do not outperform local order. For down,
the second choice causes the largest mean error. Even so, the top two
predicted by $G_\ell$ contain the worst direction in 40.5\% of cases
versus 32.1\% for local order, a gain of 8.4 percentage points
(paired 95\% interval: 5.8--10.9).

\textbf{Observation 2.} For each of the 312 linear layers, we compute
both coordinates in \cref{fig:act-weight} by dividing the four-step sum
of squared activation-only or weight-only errors by the corresponding
four-step BF16 layer output energy, then averaging the ratios over
24 prompts. Using BF16 input $X$ gives the weight-only error
$X\Delta_R^\top$, while the W4A4 weight term in
\cref{eq:error-decomposition} uses the quantized input and equals
$X_4\Delta_R^\top$.

\section{Video evaluation details}
\label{app:video-evaluation}

\begin{table*}[t]
  \captionsetup{skip=3pt}
  \caption{Additional video generation results on Wan2.1-14B.}
  \label{tab:wan21-results}
  \centering
  \footnotesize
  \renewcommand{\arraystretch}{0.9}
  \setlength{\tabcolsep}{3pt}
  \begin{tabular*}{\textwidth}{@{\extracolsep{\fill}}l*{9}{r}@{}}
    \toprule
    \multicolumn{7}{l}{} & \multicolumn{3}{c}{VBench} \\
    \cmidrule(l){8-10}
    Method & CLIPSIM$\uparrow$ & VQA(Avg)$\uparrow$ & $\Delta$FSCore$\downarrow$ & PSNR$\uparrow$ & SSIM$\uparrow$ & LPIPS$\downarrow$ & S.C.$\uparrow$ & A.Q.$\uparrow$ & I.Q.$\uparrow$ \\
    \midrule
    BF16 & 0.192 & 90.79 & -- & -- & -- & -- & 97.32 & 68.78 & 75.47 \\
    ViDiT-Q & 0.150 & 63.11 & 1.875 & 15.70 & 0.623 & 0.310 & 91.11 & 57.65 & 61.93 \\
    $\mathrm{S}^{2}$Q-VDiT & 0.162 & 86.31 & 1.024 & 17.30 & 0.685 & 0.258 & 94.79 & 66.53 & 72.65 \\
    DeltaQuant & 0.181 & 81.83 & 0.748 & 19.43 & 0.719 & 0.268 & 95.48 & 65.50 & 71.05 \\
    SVDQ & 0.163 & 70.53 & 1.647 & 16.27 & 0.615 & 0.367 & 91.70 & 60.64 & 65.00 \\
    H-SVDQ & 0.175 & 83.17 & 1.062 & 19.43 & 0.794 & 0.254 & 94.38 & 65.55 & 71.81 \\
    GPTAQ & 0.179 & 85.50 & 0.981 & 18.71 & 0.721 & 0.298 & 94.20 & 66.37 & 72.52 \\
    Ours & \textbf{0.188} & \textbf{88.63} & \textbf{0.528} & \textbf{20.72} & \textbf{0.830} & \textbf{0.210} & \textbf{96.55} & \textbf{68.04} & \textbf{74.46} \\
    \bottomrule
  \end{tabular*}
\end{table*}

For each video or image checkpoint, we use 25\% of prompts for calibration and the remaining 75\% for evaluation, with the same evaluation prompts for all methods in a comparison.
The quality tables report observed rankings on these matched panels. KLI-Quant improves fidelity across multiple backbones and sampling schedules, but a narrow lead on an individual metric does not establish a statistically significant difference.
The six quality and fidelity metrics use the full evaluation panel. \Cref{tab:video-generation-settings} lists the generation settings for all five video checkpoints.
\Cref{tab:h3-results} compares the 50-step MiniMax-H3 Base with the four-step MiniMax-H3 Fast.
MiniMax-H3 Fast uses the dense-attention, data-free DMD2 adapter~\cite{fastvideo2026fasth3dense} without classifier-free guidance.
The H3 prompts come from EvalCrafter and VBench. On MiniMax-H3 Fast, video quality and step output error are evaluated on the same prompts: each method generates videos along its own trajectory, whereas step output error is measured at shared BF16 teacher states.
To measure motion deviation ($\Delta\mathrm{FSCore}$), we use RAFT-large~\cite{teed2020raft} to estimate optical flow between consecutive frames, average its magnitude over pixels and frame pairs, and report the mean absolute difference between the resulting scores of paired quantized and BF16 videos.
For VQA(Avg), we apply the DOVER~\cite{wu2023dover} sigmoid mappings separately to the aesthetic and technical predictions, average the two scores on their 0--100 scale for each video, and then average over videos.

\begin{table}[t]
  \caption{Generation settings for video quality evaluation.}
  \label{tab:video-generation-settings}
  \centering
  \small
  \setlength{\tabcolsep}{3pt}
  \begin{tabular*}{\linewidth}{@{\extracolsep{\fill}}lrrrr@{}}
    \toprule
    Checkpoint & Steps & Resolution & Frames & FPS \\
    \midrule
    HunyuanVideo-13B & 50 & $960\times544$ & 129 & 24 \\
    Wan2.1-14B       & 50 & $832\times480$ & 81 & 16 \\
    Wan2.2-A14B      & 50 & $832\times480$ & 81 & 16 \\
    MiniMax-H3 Base  & 50 & $960\times544$ & 124 & 24 \\
    MiniMax-H3 Fast  &  4 & $1344\times768$ & 124 & 24 \\
    \bottomrule
  \end{tabular*}
\end{table}

\subsection{Supplementary reconstruction analysis}
\label{app:posthead-diagnostic}
\label{sec:analysis-results}
\label{sec:posthead-results}

\begin{figure*}[!t]
  \centering
  \captionsetup{skip=3pt}
  \begin{minipage}[t]{\columnwidth}\centering
    \includegraphics[width=\linewidth]{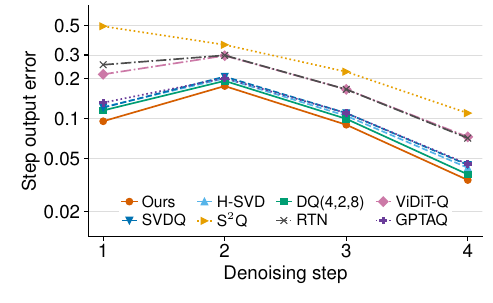}\par
    {\small\textbf{(a) Per-step output error}\par}
  \end{minipage}\hfill
  \begin{minipage}[t]{\columnwidth}\centering
    \includegraphics[width=\linewidth]{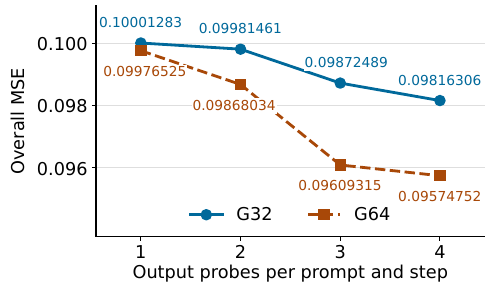}\par
    {\small\textbf{(b) Overall error by probe count}\par}
  \end{minipage}
  \caption{MiniMax-H3 Fast output MSE (lower is better). Panel (a) uses a log scale. Panel (b) shows four-step means on a linear scale.}
  \label{fig:posthead-mse}
\end{figure*}

\textbf{Metric definition and averaging.}
For method $q$, prompt $i$, and denoising step $k$, let $x^T_{i,k}$ be the shared BF16 teacher state and $f_q,f_T$ the quantized and reference video outputs of the network. With $N_{\mathrm{eval}}$ evaluation prompts, the per-prompt error, per-step mean, and overall mean are
\begin{equation}
  \begin{aligned}
    e_{i,k}^{(q)}
      &=\frac{\|f_q(x^T_{i,k})-f_T(x^T_{i,k})\|_2^2}{m_{i,k}},\\
    \bar e_k^{(q)}&=\frac1{N_{\mathrm{eval}}}\sum_{i=1}^{N_{\mathrm{eval}}} e_{i,k}^{(q)},
    \qquad \bar e^{(q)}=\frac14\sum_k\bar e_k^{(q)}.
  \end{aligned}
  \label{eq:teacher-state-mse}
\end{equation}
Here $m_{i,k}$ counts video-output elements. Prompts and steps receive equal weight.

\textbf{Configuration and comparison scope.}
\Cref{fig:posthead-mse}(a) compares KLI-Quant with the baselines at each denoising step on the H3 evaluation panel.
The rank-64 full-method checkpoint in \cref{tab:ablation} is an exception to the two-probe setting stated in \cref{sec:ablation}: it used four output probes (G64/P4). \Cref{fig:posthead-mse}(a) instead uses rank 128 and G64/P2, so the difference in four-step mean error does not isolate the effect of bypass rank.
Panel~(b) holds the bypass rank at 128 and varies the number of output probes from one to four at each token quota (G32 and G64). In G$n$/P$p$, $n$ is the total number of sensitivity-token rows sampled across all parallel ranks per calibration prompt, denoising step, and layer or shared projection group. The $p$ output probes per prompt and step estimate downstream sensitivity.
SVDQuant uses weight-SVD allocation and one-pass $H_X$ reconstruction. Our NVFP4 adaptation of H-SVDQ~\cite{wang2026hsvdquant} initializes the bypass using input-weighted SVD, quantizes the residual using the deflated Hessian, and then refits the output factor of the bypass in closed form. We use these three steps in both the main-paper quality comparisons and \cref{fig:posthead-mse}(a), where the legend abbreviates H-SVDQ as H-SVD.
We adapt the official ViDiT-Q implementation~\cite{zhao2025viditq} to the common NVFP4 format, retaining channel balancing and per-token activation scaling.
$\mathrm{S}^{2}$Q denotes our adaptation of $\mathrm{S}^{2}$Q-VDiT~\cite{feng2025s2qvdit}, developed without publicly released reference code. It selects calibration examples using the Hessian and distills teacher outputs on attention-selected tokens. DQ(4,2,8) denotes DeltaQuant with a fixed $4\times2\times8$ spatiotemporal cube.
The GPTAQ and SVDQuant baselines share the same fixed bypass obtained by plain weight SVD. For GPTAQ, we adapt the official implementation~\cite{li2025gptaq} to calibrate the remaining residual.

\begin{table}[t]
  \caption{G64/P2 prompt win rates (\%). For each prompt, $\delta_k$ is the MSE of G64/P2 minus the baseline MSE at step $k$. Both columns use all evaluation prompts as the denominator.}
  \label{tab:mse-prompt-wins}
  \centering
  \small
  \setlength{\tabcolsep}{3pt}
  \begin{tabular*}{\linewidth}{@{\extracolsep{\fill}}lrr@{}}
    \toprule
    Baseline & \shortstack{Win on average\\$\tfrac14\sum_{k=1}^4\delta_k<0$} & \shortstack{Win at every step\\$\delta_k<0$ for all $k$} \\
    \midrule
    DeltaQuant & 76.0 & 39.0 \\
    SVDQ & 90.0 & 64.0 \\
    RTN & 99.0 & 95.5 \\
    \bottomrule
  \end{tabular*}
\end{table}

\textbf{Step output error.}
G64/P2 has lower mean MSE than the compared baselines at each step (\cref{fig:posthead-mse}(a)). Its error peaks at step~2, is lowest at step~4, and averages 0.09868, 11.08\% below DeltaQuant. In panel~(b), MSE falls with each additional probe at both quotas. From one to four probes, it falls by 1.85\% at G32 and 4.03\% at G64, with G64/P4 lowest (0.09575). Against DeltaQuant, G64/P2 improves the four-step mean on 76\% of prompts: it wins every step on 39\% and ties or loses at least one step on the other 37\% (\cref{tab:mse-prompt-wins}). These rates describe variation across prompts, not calibration runs.

\subsection{Allocation-only rank and probe study}
\label{app:rank-probe-allocation}

We evaluate bypass allocation at denoising step~2 on the same held-out sources. At eight ranks from 1 to 128, we compare unweighted SVD of the weights with task-sensitive allocation using 2, 4, 8, 16, or 24 probes. Across the five probe settings, larger ranks generally yield lower mean step output error, but more probes at a fixed rank do not consistently reduce it (\cref{fig:rank-probe-allocation}). At rank~128, an additional control sets $G=I$ while retaining the activation-error metric $H_{\Delta_X}$. \Cref{tab:rank-probe-128} reports mean error, its reduction relative to unweighted SVD, and the percentage of sources with lower error than SVD. The $G=I$ control improves on SVD, and including downstream sensitivity further reduces mean error at all five probe counts. For 24 probes versus SVD, the 95\% paired bootstrap interval for the mean error difference across sources is $[-0.141342,-0.089129]$.

\begin{table}[t]
  \caption{Rank-128 allocation results at step 2.}
  \label{tab:rank-probe-128}
  \centering
  \small
  \setlength{\tabcolsep}{3pt}
  \begin{tabular*}{\linewidth}{@{\extracolsep{\fill}}lrrr@{}}
    \toprule
    Allocation & Step Output Error $\downarrow$ & Reduction & Wins (\%) \\
    \midrule
    Unweighted SVD & 0.520756 & --- & --- \\
    $G=I$ & 0.489866 & 5.93\% & 73.44 \\
    Task-sensitive P2 & 0.429432 & 17.54\% & 90.63 \\
    Task-sensitive P4 & 0.422235 & 18.92\% & 92.19 \\
    Task-sensitive P8 & 0.416047 & 20.11\% & 95.31 \\
    Task-sensitive P16 & 0.423049 & 18.76\% & 93.75 \\
    Task-sensitive P24 & \textbf{0.408275} & \textbf{21.60\%} & \textbf{98.44} \\
    \bottomrule
  \end{tabular*}
\end{table}

\subsection{Sensitivity in allocation and calibration}
\label{app:stage2-sensitivity}

ARHQ~\cite{wang2026arhq} selects a bypass to minimize the transmitted activation error, $\|R H_{\Delta_X}^{1/2}\|_F^2$. This criterion, used by the input-only control in \cref{tab:stage2-metric}, omits residual weight error. ARHQ-w uses that error to refine the bypass. KLI-Quant instead selects the bypass for activation error weighted by downstream sensitivity $G$ and gives residual weight error a separate calibration objective.

With the stored bypass fixed, JSC selects codes and scales for $R=W-BA$ under $H_{X_4}$. MSE uses shared BF16 teacher states over four steps (\cref{eq:teacher-state-mse}). SSIM and LPIPS use paired BF16 videos.

\par\smallskip\noindent
\begin{minipage}{\linewidth}
  \centering
  \captionsetup{type=table,skip=3pt}
  \caption{Downstream weighting on rank-64 MiniMax-H3 Fast. All allocations use $H_{\Delta_X}$. The JSC column gives the output metric for calibration. The first and last rows are from \cref{tab:ablation}.}
  \label{tab:stage2-metric}
  \footnotesize
  \setlength{\tabcolsep}{2pt}
  \begin{tabular*}{\linewidth}{@{\extracolsep{\fill}}lcrrr@{}}
    \toprule
    Bypass allocation & JSC & \shortstack[r]{Step output\\MSE$\downarrow$} & SSIM$\uparrow$ & LPIPS$\downarrow$ \\
    \midrule
    Input-only ($I$)     & $I$ & 0.1159  & 0.4583 & 0.531 \\
    Task-sensitive ($G$) & $I$ & 0.09815 & \textbf{0.4653} & 0.537 \\
    \addlinespace[2pt]
    Input-only ($I$)     & $G$ & 0.1137  & 0.4629 & 0.541 \\
    Task-sensitive ($G$) & $G$ & \textbf{0.09736} & 0.4640 & \textbf{0.524} \\
    \bottomrule
  \end{tabular*}
\end{minipage}
\par\smallskip

Even with full-$G$ JSC in both arms (last two rows), using downstream sensitivity to select the bypass lowers mean output MSE by 14.37\% and LPIPS by 3.14\% relative to input-only allocation on this panel. The last row reproduces the rank-64 G64/P4 full-method result from \cref{tab:ablation}. The second row reuses its stored bypass and changes only the JSC calibration metric. Using $G$ rather than $I$ lowers MSE by 0.80\% and LPIPS by 2.42\%, while SSIM decreases from 0.4653 to 0.4640.

\section{Image evaluation details}
\label{app:image-evaluation}

\begin{table*}[t]
  \captionsetup{skip=3pt}
  \caption{Image generation results on Qwen-Image with 50 steps.}
  \label{tab:qwen-image-base50}
  \centering
  \footnotesize
  \renewcommand{\arraystretch}{0.9}
  \setlength{\tabcolsep}{4pt}
  \begin{tabular}{@{}l@{\hspace{8pt}}rrrr@{\hspace{10pt}}rrrr@{\hspace{10pt}}rrr@{}}
    \toprule
    & \multicolumn{4}{c}{MJHQ} & \multicolumn{4}{c}{sDCI} & \multicolumn{3}{c}{LongText-Bench} \\
    \cmidrule(lr){2-5}\cmidrule(lr){6-9}\cmidrule(l){10-12}
    Method & PSNR$\uparrow$ & LPIPS$\downarrow$ & CLIP-T$\uparrow$ & Reward$\uparrow$ & PSNR$\uparrow$ & LPIPS$\downarrow$ & CLIP-T$\uparrow$ & Reward$\uparrow$ & PSNR$\uparrow$ & LPIPS$\downarrow$ & Acc.$\uparrow$ \\
    \midrule
    BF16 & -- & -- & 29.483 & 1.354 & -- & -- & 26.676 & 1.254 & -- & -- & 93.508 \\
    RTN & 18.27 & 0.327 & 29.039 & 1.306 & 17.68 & 0.356 & 26.605 & 1.190 & 14.90 & 0.334 & 83.027 \\
    ViDiT-Q & 18.34 & 0.327 & 29.085 & 1.308 & 17.77 & 0.318 & 26.651 & 1.204 & 14.92 & 0.335 & 84.090 \\
    $\mathrm{S}^{2}$Q-VDiT & 14.38 & 0.452 & 29.057 & 1.338 & 13.99 & 0.519 & 27.280 & 1.230 & 13.62 & 0.461 & 85.955 \\
    SVDQ & 19.37 & 0.241 & 29.457 & 1.363 & 18.54 & 0.279 & \textbf{27.767} & \textbf{1.293} & 16.03 & 0.242 & 92.027 \\
    H-SVDQ & 19.48 & 0.243 & 29.149 & 1.332 & 17.94 & 0.271 & 27.481 & 1.258 & 15.99 & 0.286 & 93.109 \\
    GPTAQ & 19.10 & 0.263 & 29.067 & 1.320 & 18.36 & 0.291 & 27.108 & 1.215 & 15.86 & 0.295 & 92.937 \\
    Ours & \textbf{20.75} & \textbf{0.216} & \textbf{29.663} & \textbf{1.377} & \textbf{18.96} & \textbf{0.231} & 27.198 & 1.285 & \textbf{16.32} & \textbf{0.233} & \textbf{93.491} \\
    \bottomrule
  \end{tabular}
\end{table*}

\textbf{Checkpoints and pairing.}
The 50-step base and 4-step distilled Qwen-Image models (\cref{sec:evaluation-scope}) generate $1024\times1024$ images with classifier-free guidance scales of 4.0 and 1.0, respectively. Each quantized image is paired with a BF16 image from the same checkpoint, prompt, and initial noise.

\textbf{Datasets and calibration.}
Evaluation prompts come from MJHQ-5000, shortened sDCI captions, and English LongText-Bench. Each method generates one image per prompt on MJHQ and sDCI and four on LongText-Bench. Calibration uses equal numbers of MJHQ, sDCI, and LongText-Bench prompts.

\textbf{Image results.}
PSNR and LPIPS use the paired BF16 image as reference. CLIP-T measures text--image alignment, and ImageReward~\cite{xu2023imagereward} measures learned preference. At 50 steps (\cref{tab:qwen-image-base50}), KLI-Quant leads the quantized methods in nine of eleven comparisons: all four MJHQ metrics, sDCI PSNR and LPIPS, and all three LongText-Bench metrics. SVDQ leads on sDCI CLIP-T and ImageReward. KLI-Quant has the highest English TextAcc among the quantized methods at both step counts (\cref{tab:qwen-image-results,tab:qwen-image-base50}).

\textbf{English text recognition.}
Following the English LongText-Bench protocol~\cite{xomni2025longtext}, we use \texttt{Qwen2.5-VL-7B-Instruct} to read the text in each generated image. TextAcc is the percentage of reference word occurrences recovered across all prompts and all four generations per prompt, counting the reference once per generation. Each recognized word can match one reference occurrence of the same word. Word order and extra words do not affect the score.

\section{Language-model evaluation}
\label{app:llm-results}

\begin{table*}[t]
  \captionsetup{skip=3pt}
  \caption{Language-model results on Qwen3-8B. Accuracies are in \%.}
  \label{tab:llm-results}
  \centering
  \footnotesize
  \renewcommand{\arraystretch}{0.9}
  \setlength{\tabcolsep}{3pt}
  \begin{tabular*}{\textwidth}{@{\extracolsep{\fill}}l*{9}{r}@{}}
    \toprule
    & \multicolumn{2}{c}{Perplexity} & \multicolumn{7}{c}{Accuracy} \\
    \cmidrule(lr){2-3}\cmidrule(l){4-10}
    Method & WikiText-2$\downarrow$ & C4$\downarrow$ & AIME 2025$\uparrow$ & MMLU-Pro$\uparrow$ & HellaSwag$\uparrow$ & ARC-C$\uparrow$ & PIQA$\uparrow$ & WinoGrande$\uparrow$ & BoolQ$\uparrow$ \\
    \midrule
    BF16 & 9.730 & 13.103 & 22.50 & 74.94 & 74.90 & 56.31 & 77.42 & 67.80 & 86.51 \\
    RTN & 10.073 & 13.641 & 21.67 & 61.19 & 73.28 & 53.41 & 76.22 & 67.25 & 85.23 \\
    SmoothQuant & 10.089 & 13.652 & 21.73 & 61.09 & 73.14 & 54.36 & 75.83 & 67.03 & 85.44 \\
    SVDQ & 9.743 & 13.256 & 22.05 & 64.16 & 73.67 & 56.01 & 76.72 & \textbf{67.53} & 85.74 \\
    Ours & \textbf{9.732} & \textbf{13.15} & \textbf{22.32} & \textbf{72.53} & \textbf{74.36} & \textbf{56.21} & \textbf{76.93} & 67.42 & \textbf{86.29} \\
    \bottomrule
  \end{tabular*}
\end{table*}

\textbf{Setup.}
\Cref{tab:llm-results} applies the methods to the language model Qwen3-8B~\cite{yang2025qwen3}. All methods use the NVFP4 weight and activation format of the diffusion experiments. SmoothQuant~\cite{xiao2023smoothquant} moves activation outliers into the weights before NVFP4 quantization. SVDQ uses a rank-64 weight-SVD bypass with GPTQ residual calibration, as in \cref{sec:evaluation-scope}, and KLI-Quant also uses rank-64 bypasses. We report perplexity on WikiText-2~\cite{merity2017pointer} and C4~\cite{raffel2020t5}, and accuracy on AIME 2025, MMLU-Pro~\cite{wang2024mmlupro}, HellaSwag~\cite{zellers2019hellaswag}, ARC-Challenge~\cite{clark2018arc}, PIQA~\cite{bisk2020piqa}, WinoGrande~\cite{sakaguchi2020winogrande}, and BoolQ~\cite{clark2019boolq}.

\textbf{Results.}
KLI-Quant is best among quantized methods on eight of the nine metrics. Its perplexity is 0.002 above BF16 on WikiText-2 and 0.047 above on C4, compared with 0.343 and 0.538 for RTN. MMLU-Pro separates the methods most: relative to BF16, RTN and SmoothQuant lose 13.75 and 13.85 points, SVDQ loses 10.78, and KLI-Quant loses 2.41. On HellaSwag, ARC-Challenge, PIQA, WinoGrande, and BoolQ, KLI-Quant is within 0.54 points of BF16, and SVDQ is 0.11 points higher than KLI-Quant on WinoGrande. On AIME 2025, every quantized method is within 0.83 points of BF16, and KLI-Quant is 0.18 points below it.

\section{Additional qualitative comparisons}
\label{app:qualitative-video}

Six full video comparisons and two image grids complement the main-text examples in \cref{fig:h3-dist-main-qualitative,fig:h3-dist-qualitative,fig:qwen-dist-main-qualitative}. These selected examples illustrate appearance and composition rather than statistical representativeness.

\textbf{Video comparisons.}
The three MiniMax-H3 base examples use 50 requested steps and precede the three MiniMax-H3 Fast examples, which use four steps. Columns show frames 1, 32, 63, 93, and 124. Rows show BF16, RTN, ViDiT-Q, S2Q-VDiT, DeltaQuant ($4\times2\times8$), SVDQ, H-SVDQ, GPTAQ, and Ours. Snapshots do not establish continuous-motion quality, and captions describe the requested scenes rather than certify that every requested element is reproduced.

\textbf{Image comparisons.}
\Cref{fig:qwen-dist-main-qualitative} in the main text shows one 4-step example per dataset. The full 50-step and 4-step Qwen-Image grids in \cref{fig:qwen-base-qualitative,fig:qwen-dist-qualitative} provide three examples per dataset from MJHQ, sDCI, and English LongText-Bench. The images illustrate scene composition, object appearance, and text layout.

\begin{figure*}[!tp]
  \centering
  \input{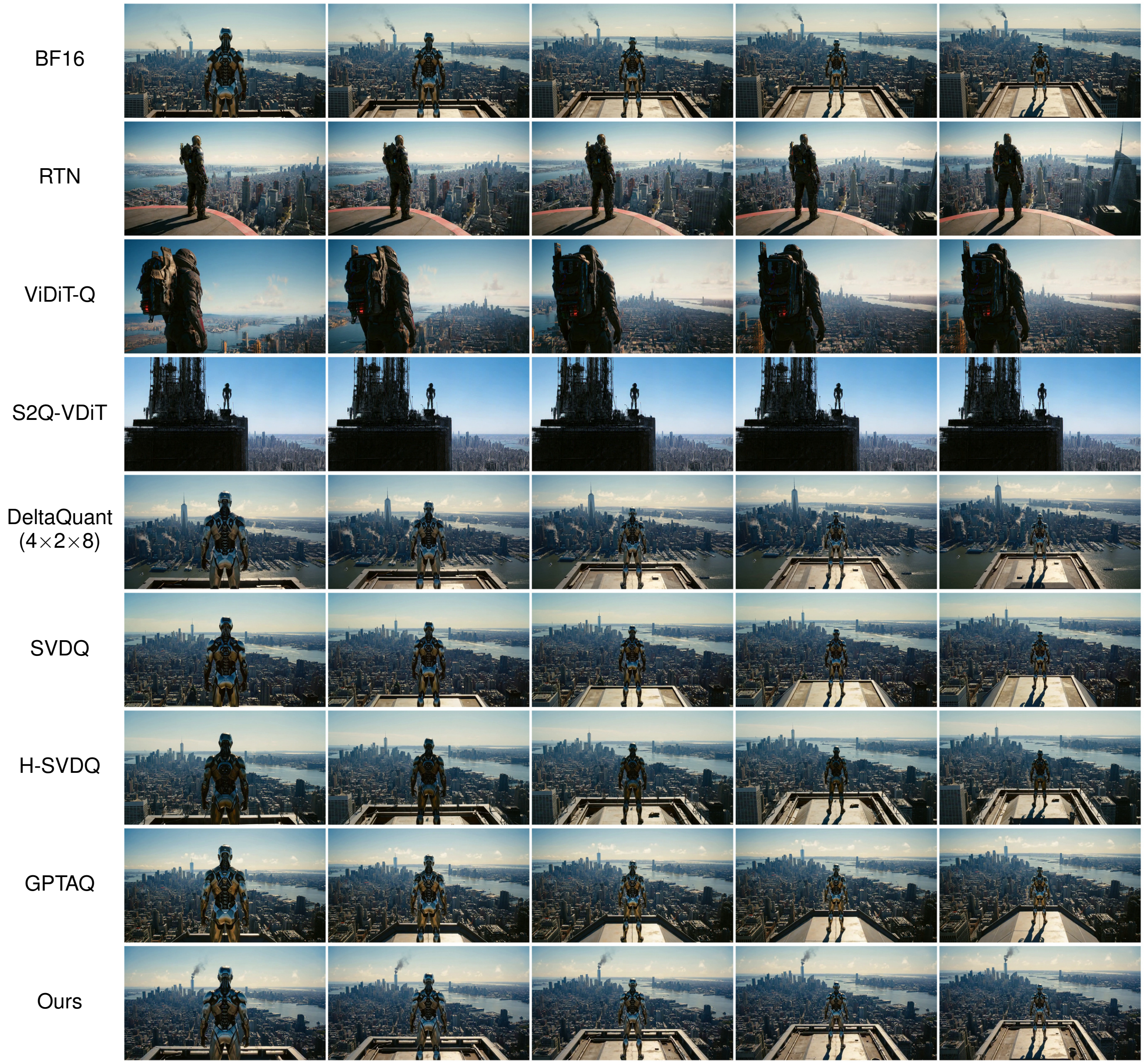}
  \caption{50-step MiniMax-H3 base: a cyborg viewed from behind surveys a futuristic city from a skyscraper.}
  \label{fig:h3-base-qualitative-0003}
\end{figure*}

\begin{figure*}[p]
  \centering
  \input{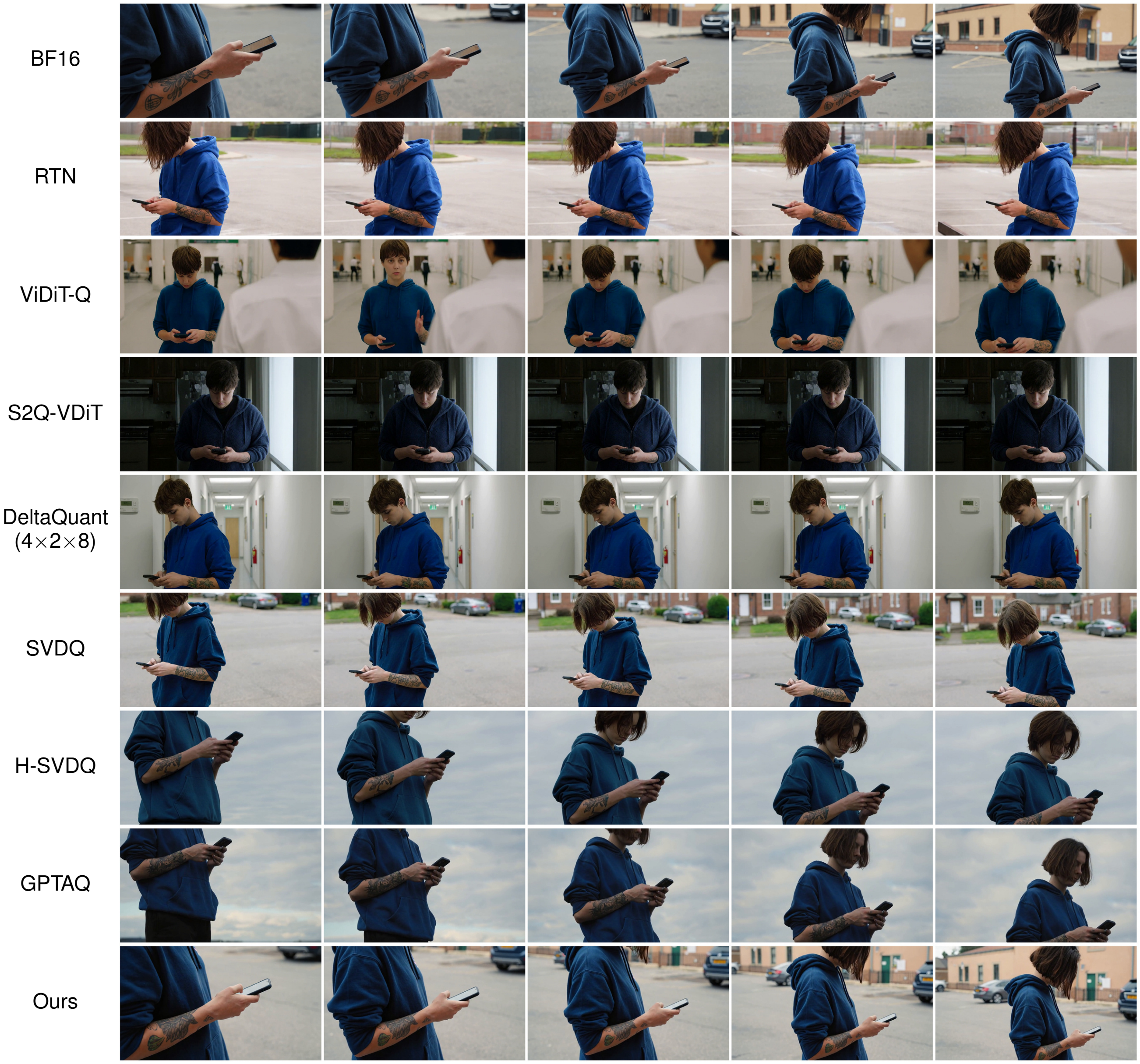}
  \caption{50-step MiniMax-H3 base: a young adult with short brown hair, a forearm tattoo, and a blue hoodie checks a phone.}
  \label{fig:h3-base-qualitative-0004}
\end{figure*}

\begin{figure*}[p]
  \centering
  \input{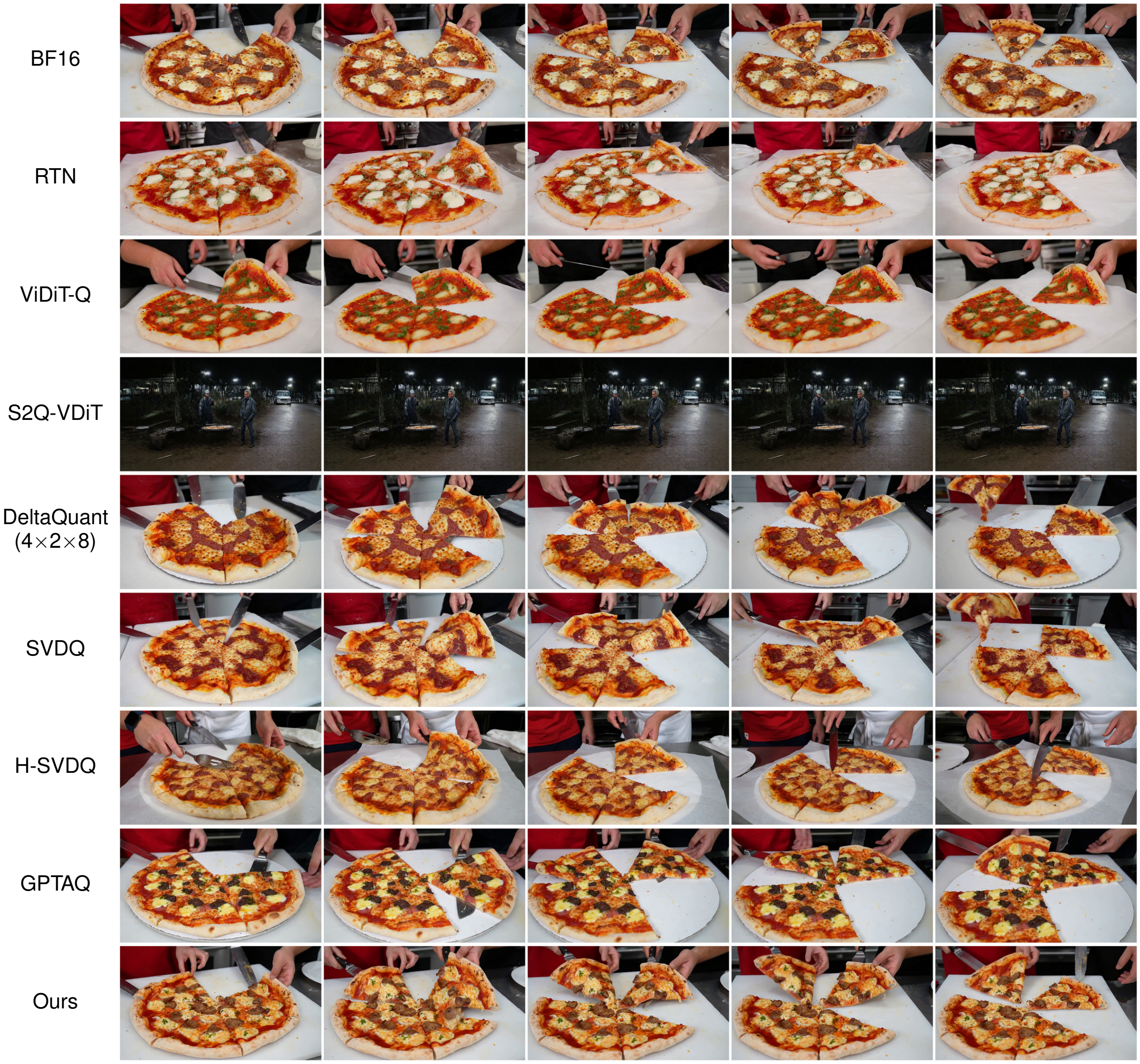}
  \caption{50-step MiniMax-H3 base: one bear and two people make pizza.}
  \label{fig:h3-base-qualitative-0012}
\end{figure*}

\begin{figure*}[p]
  \centering
  \input{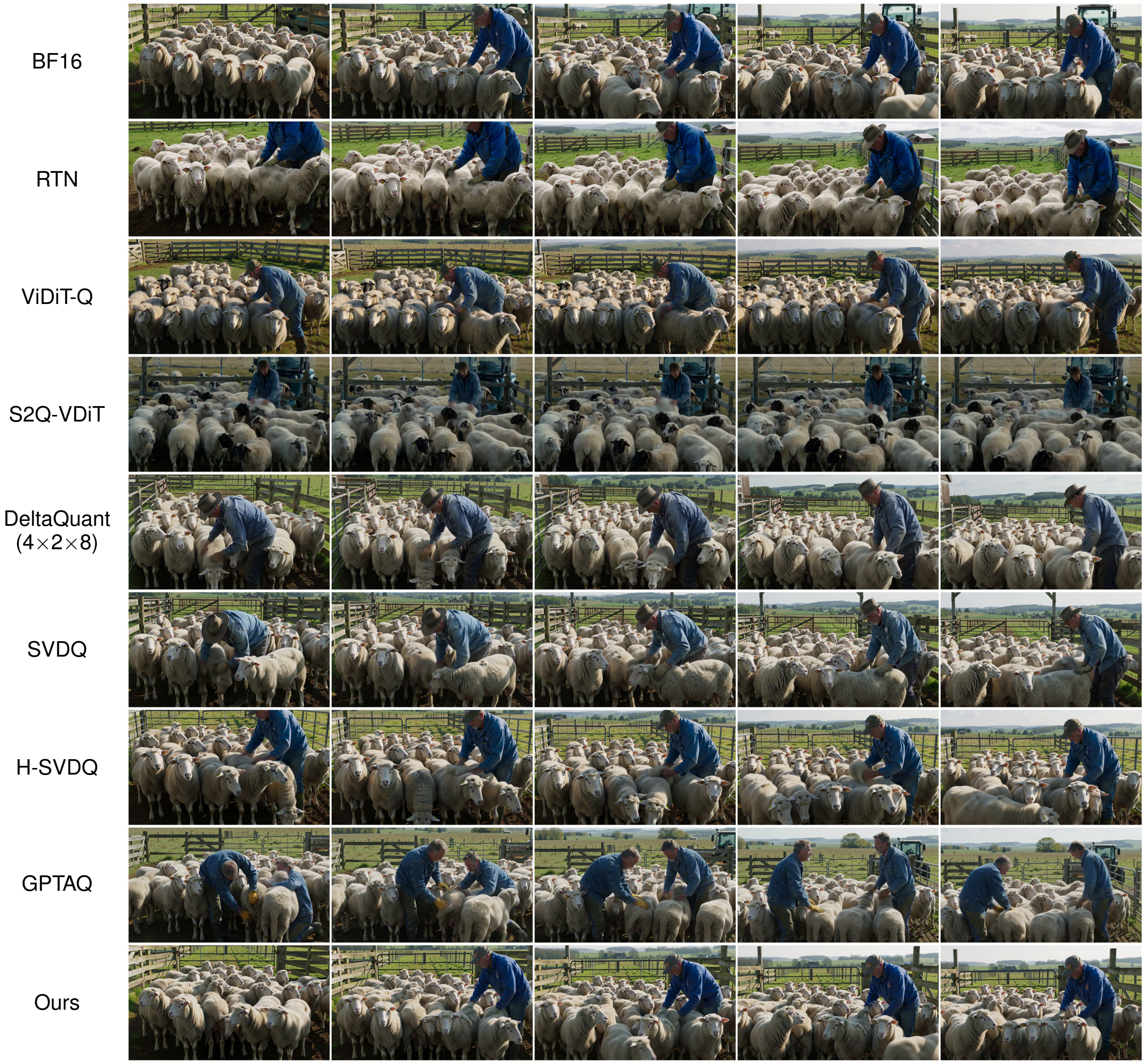}
  \caption{Four-step MiniMax-H3 Fast: a farmer shears and counts a flock of twenty sheep.}
  \label{fig:h3-dist-qualitative-0014}
\end{figure*}

\begin{figure*}[p]
  \centering
  \input{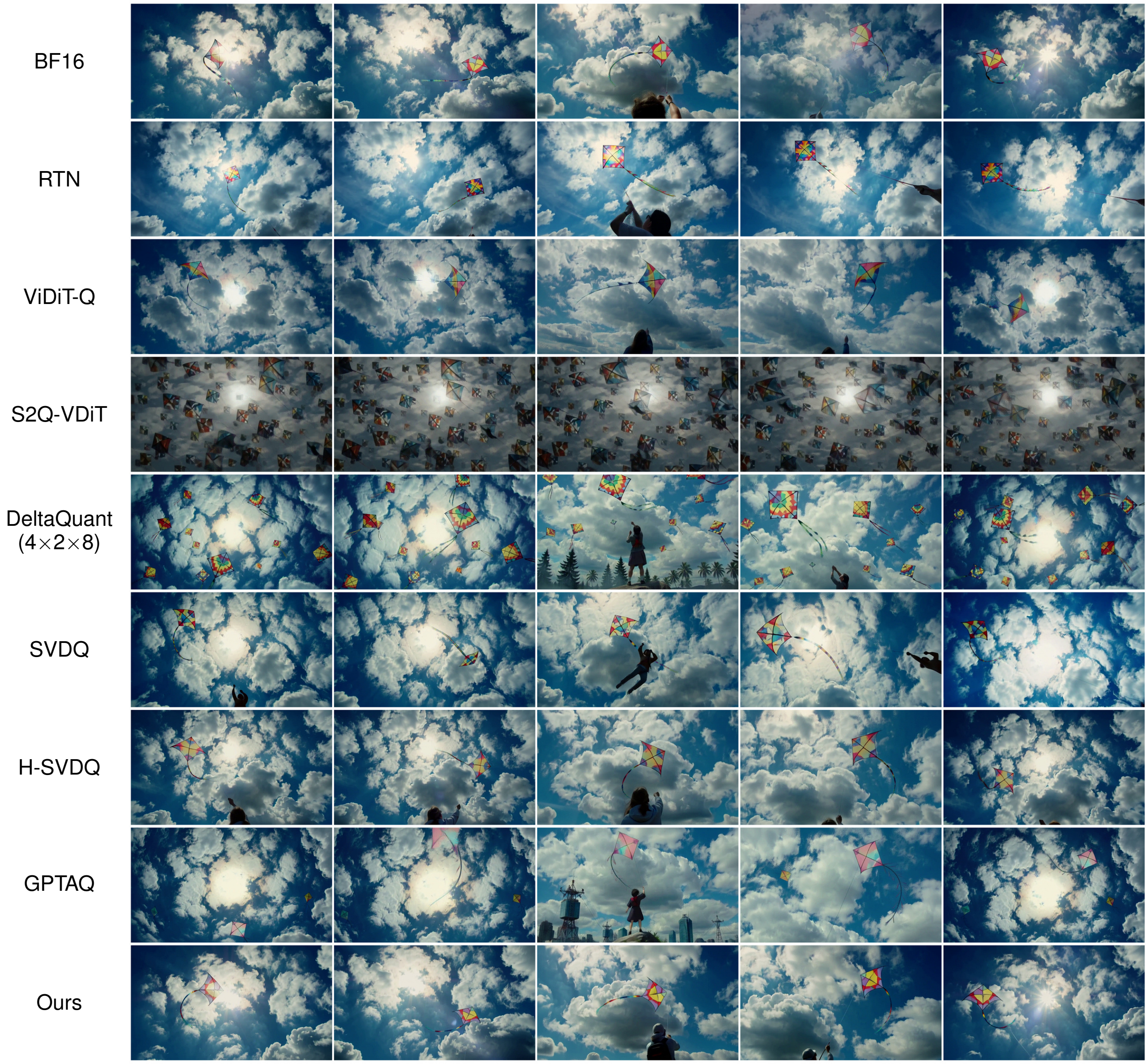}
  \caption{Four-step MiniMax-H3 Fast: a person flies a kite against a blue background with a central white light.}
  \label{fig:h3-dist-qualitative-0041}
\end{figure*}

\begin{figure*}[p]
  \centering
  \input{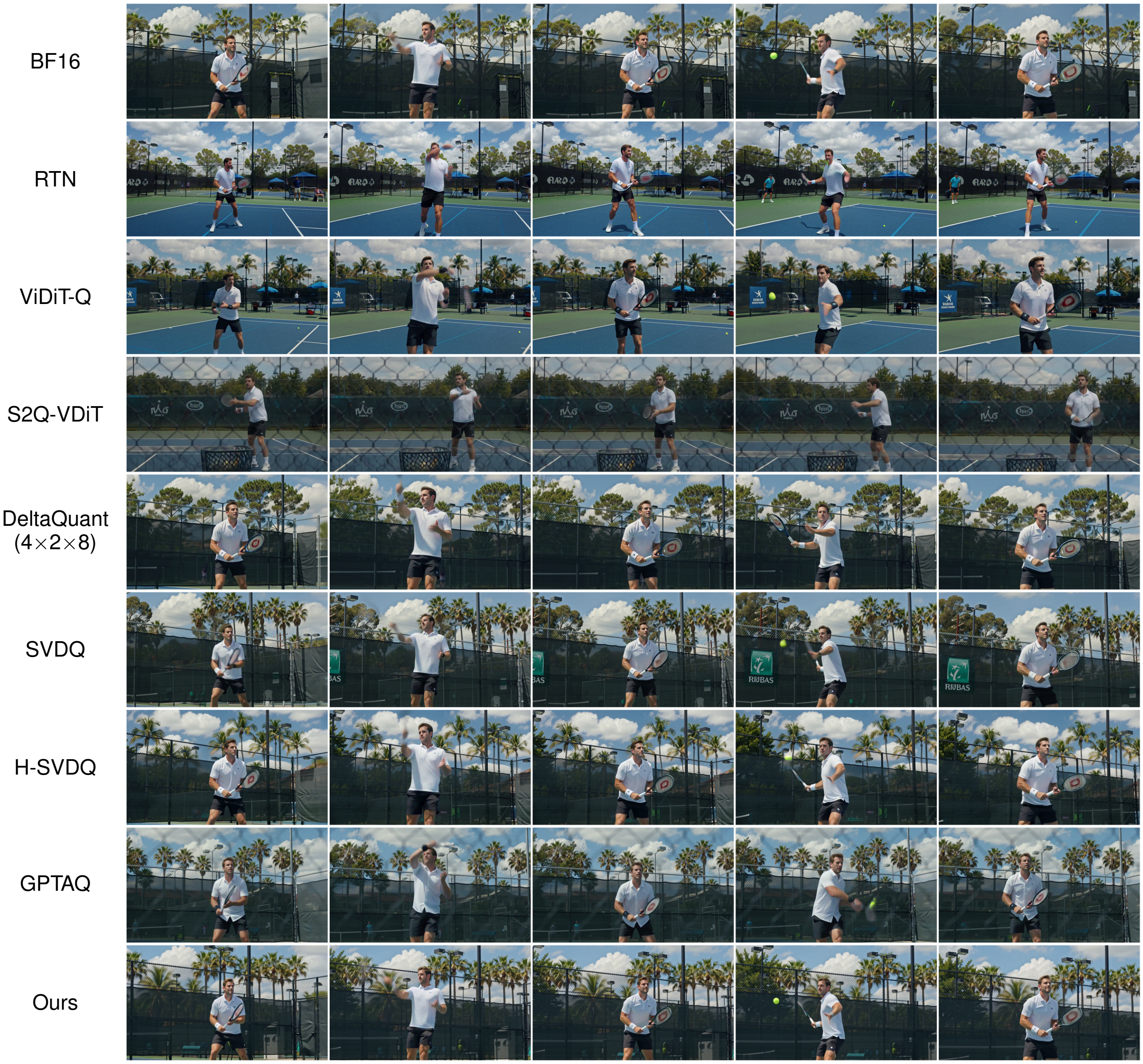}
  \caption{Four-step MiniMax-H3 Fast: a man in a white shirt and black shorts hits a green tennis ball beneath a blue sky.}
  \label{fig:h3-dist-qualitative-0046}
\end{figure*}

\begin{figure*}[p]
  \centering
  \input{artwork/labels/qwen-base}
  \caption{Qwen-Image, 50 steps. The three groups show MJHQ, sDCI, and English LongText-Bench, with three examples per dataset. Each row uses a common prompt and seed.}
  \label{fig:qwen-base-qualitative}
\end{figure*}

\begin{figure*}[p]
  \centering
  \input{artwork/labels/qwen-dist}
  \caption{Qwen-Image, 4 steps. The three groups show MJHQ, sDCI, and English LongText-Bench, with three examples per dataset. Each row uses a common prompt and seed.}
  \label{fig:qwen-dist-qualitative}
\end{figure*}

\section{Latency analysis}
\label{app:latency-details}

\setlength{\dblfloatsep}{7pt}
\setlength{\dbltextfloatsep}{7pt}
\subsection{GEMM and end-to-end latency}
\label{sec:latency-results}

\begin{figure*}[p]
  \begin{minipage}{\textwidth}
  \captionof{table}{Median end-to-end latency.}
  \label{tab:latency-e2e}
  \centering
  \small
  \renewcommand{\arraystretch}{0.95}
  \setlength{\tabcolsep}{3pt}
  \begin{tabular*}{\textwidth}{@{\extracolsep{\fill}}lrrrrrrrr@{}}
    \toprule
    Model & Devices & A (s) & B (s) & C (s) & A$\to$B & B$\to$C & A$\to$C & CV (\%) \\
    \midrule
    Wan2.1-1.3B       & 1 & 150.17  & 108.50 & 83.95  & $1.38\times$ & $1.29\times$ & $1.79\times$ & $\le0.22$ \\
    Qwen-Image       & 1 & 44.46   & 42.29  & 17.30  & $1.05\times$ & $2.45\times$ & $2.57\times$ & $\le0.44$ \\
    Wan2.1-14B       & 4 & 981.74  & 719.94 & 503.89 & $1.36\times$ & $1.43\times$ & $1.95\times$ & $\le0.04$ \\
    Wan2.1-14B       & 8 & 564.82  & 408.62 & 301.14 & $1.38\times$ & $1.36\times$ & $1.88\times$ & $\le0.01$ \\
    HunyuanVideo-13B & 4 & 1013.19 & 659.69 & 551.12 & $1.54\times$ & $1.20\times$ & $1.84\times$ & $\le0.01$ \\
    HunyuanVideo-13B & 8 & 721.82  & 462.94 & 396.88 & $1.56\times$ & $1.17\times$ & $1.82\times$ & $\le0.03$ \\
    Wan2.2-A14B$^\dagger$ & 4 & 798.49 & 589.93 & 405.26 & $1.35\times$ & $1.46\times$ & $1.97\times$ & $\le0.02$ \\
    Wan2.2-A14B$^\dagger$ & 8 & 480.68 & 355.58 & 242.52 & $1.35\times$ & $1.47\times$ & $1.98\times$ & $\le0.01$ \\
    MiniMax-H3$^\dagger$  & 4 & 51.30  & 43.06  & 26.64  & $1.19\times$ & $1.62\times$ & $1.93\times$ & $\le0.12$ \\
    MiniMax-H3$^\dagger$  & 8 & 31.03  & 27.59  & 15.92  & $1.12\times$ & $1.73\times$ & $1.95\times$ & $\le0.08$ \\
    \bottomrule
  \end{tabular*}
  \end{minipage}
  \par\smallskip
  \begin{minipage}{\textwidth}
  \captionof{table}{GEMM latency.}
  \label{tab:latency-gemm}
  \centering
  \small
  \renewcommand{\arraystretch}{0.95}
  \setlength{\tabcolsep}{3pt}
  \begin{tabular*}{\textwidth}{@{\extracolsep{\fill}}rrrrrrrrrr@{}}
    \toprule
    \multicolumn{3}{c}{Matrix shape} & \multicolumn{4}{c}{Latency (ms)} & \multicolumn{3}{c}{Latency ratio} \\
    \cmidrule(lr){1-3}\cmidrule(lr){4-7}\cmidrule(lr){8-10}
    $M$ & $K$ & $N$ & BF16 & Fused (F) & CUTLASS (CT) & cuBLASLt (LT) & F/CT & F/LT & CT/LT \\
    \midrule
    57600  & 4096  & 4096  & 1.016 & 0.402 & 0.397 & 0.375 & 1.01 & 1.07 & 1.06 \\
    57600  & 4096  & 11008 & 2.748 & 0.888 & 0.872 & 0.855 & 1.02 & 1.04 & 1.02 \\
    57600  & 11008 & 4096  & 2.937 & 0.810 & 0.730 & 0.693 & 1.11 & 1.17 & 1.05 \\
    165968 & 4096  & 4096  & 2.991 & 0.964 & 0.936 & 0.904 & 1.03 & 1.07 & 1.04 \\
    165968 & 4096  & 11008 & 7.961 & 2.421 & 2.340 & 2.320 & 1.03 & 1.04 & 1.01 \\
    165968 & 11008 & 4096  & 8.534 & 2.222 & 1.958 & 1.881 & 1.13 & 1.18 & 1.04 \\
    14400  & 6144  & 16384 & 1.542 & 0.531 & 0.485 & 0.460 & 1.09 & 1.15 & 1.05 \\
    57600  & 6144  & 4096  & 1.550 & 0.515 & 0.476 & 0.439 & 1.08 & 1.17 & 1.08 \\
    57600  & 6144  & 16384 & 6.126 & 1.893 & 1.662 & 1.523 & 1.14 & 1.24 & 1.09 \\
    14400  & 16384 & 6144  & 1.677 & 0.518 & 0.465 & 0.448 & 1.11 & 1.16 & 1.04 \\
    57600  & 4096  & 6144  & 1.535 & 0.533 & 0.531 & 0.506 & 1.00 & 1.05 & 1.05 \\
    57600  & 16384 & 6144  & 6.298 & 1.835 & 1.593 & 1.488 & 1.15 & 1.23 & 1.07 \\
    14400  & 6144  & 6144  & 0.584 & 0.265 & 0.254 & 0.250 & 1.04 & 1.06 & 1.02 \\
    57600  & 6144  & 6144  & 2.323 & 0.719 & 0.664 & 0.611 & 1.08 & 1.18 & 1.09 \\
    \bottomrule
  \end{tabular*}
\end{minipage}

  \par\smallskip
  \input{sec/5_figure_latency}
\end{figure*}

\textbf{Comparison design.}
To evaluate the fused two-branch backend (\cref{fig:fused-kernel}), we time two successive changes to BF16 inference: first the attention implementation, then the linear-layer backend. These timing runs are separate from the quality evaluations. All MiniMax-H3 latency measurements use an 8-step distilled checkpoint, whereas Table~\ref{tab:h3-results} reports video quality for the 50-step base model and the 4-step distilled model (settings in Table~\ref{tab:video-generation-settings}). In the end-to-end measurements, group A uses dense attention, while groups B and C both use SageAttention2. A and B retain BF16 linear layers. C uses the fused NVFP4 residual and BF16 rank-up kernel for transformer-block linear layers. \Cref{tab:latency-e2e} reports median latency for each group and the end-to-end speedups for attention alone (A$\to$B), the switch to the fused backend at fixed attention (B$\to$C), and both changes together (A$\to$C). \Cref{fig:latency-normalized} plots the same latencies relative to A.

\textbf{End-to-end performance.}
Across configurations, the combined speedup (A$\to$C) is $1.79$--$2.57\times$. With attention fixed (B$\to$C), the reported speedup is $1.17$--$2.45\times$. For example, Qwen-Image latency decreases from 42.29\,s in B to 17.30\,s in C. For the four-device MiniMax-H3 configuration, the main-paper speedup is A$\to$C, $51.30/26.64\approx1.93\times$. With attention fixed, B$\to$C is $43.06/26.64\approx1.62\times$. The CV column of \cref{tab:latency-e2e} bounds the coefficient of variation across repeated runs. MiniMax-H3 and Wan2.2 additionally change from sharded parameters in A/B to a resident, unsharded transformer in C, so their speedups include this placement change ($\dagger$ in \cref{tab:latency-e2e,tab:latency-phases}). An unfused NVFP4 control was not timed, so these results do not isolate the benefit of fusion. Phase timings, BF16 scaling, and language-model timings appear in \cref{app:latency-phases,app:latency-llm}.

\textbf{GEMM performance.}
We also time the linear operation separately from attention (\cref{tab:latency-gemm}). Across 14 matrix shapes, the fused kernel achieves $2.20$--$3.84\times$ speedup over BF16, with an arithmetic mean of $3.12\times$ over per-shape speedups. With rank-64 BF16 correction included, its reported latency is $1.00$--$1.15\times$ that of tuned CUTLASS and $1.04$--$1.24\times$ that of cuBLASLt. These NVFP4 references omit the correction and use different kernels, so the ratios do not isolate bypass overhead.

\subsection{Diffusion phase timings and scaling}
\label{app:latency-phases}

The groups follow \cref{sec:latency-results}. In C, only linear layers within diffusion-transformer blocks use the NVFP4 two-branch backend. Normalization, modulation, embeddings, and time, conditioning, patch, and final projections remain in BF16. MiniMax-H3 and Wan2.2 use an unsharded, resident transformer in C and sharded parameters in A/B. Their timing differences therefore include both backend and placement effects.

\Cref{tab:latency-phases} separates denoising from VAE decoding. Denoising dominates the reported workloads, but the unchanged decoding component limits the end-to-end benefit. Per-step entries are separately aggregated step timings rather than total denoising time divided by an assumed step count. The recorded text-encoding time is approximately zero except for 0.03\,s on MiniMax-H3. Independently summarized phase timings need not sum exactly to the end-to-end latency, and the listed phases do not account for every reported overhead.

\Cref{tab:latency-scaling} reports scaling of the BF16 reference A, not of the fused backend. Increasing the device count from four to eight yields $1.40$--$1.74\times$ speedup, below the ideal $2\times$. These measurements do not establish the scaling efficiency of C or identify the relative contributions of communication and serial work.

\subsection{Language-model serving latency}
\label{app:latency-llm}

\Cref{tab:latency-llm} gives a separate single-device Qwen3-8B timing study with 1,024 input tokens and 512 output tokens per request. Group C replaces 144 transformer-block linear layers with the low-rank NVFP4 backend, using static activation scales and retaining BF16 embeddings and the output head. Concurrency is the number of simultaneous requests. TTFT denotes time to first token, and TPOT denotes time per output token.

Output throughput increases by $1.98\times$, $1.95\times$, and $1.65\times$ at concurrency 1, 8, and 64, respectively. Reduced gains at higher concurrency are consistent with a changing balance between weight traffic and computation, but these timings alone do not establish the bottleneck. This is a latency-only test using random low-rank factors and direct quantization, not a calibrated language-quality evaluation. It provides no evidence of language-model accuracy, which is evaluated separately in \cref{app:llm-results}, and no plain NVFP4 arm was measured.

\begin{table*}[p]
  \caption{Median phase timings.}
  \label{tab:latency-phases}
  \centering
  \small
  \setlength{\tabcolsep}{3pt}
  \begin{tabular*}{\textwidth}{@{\extracolsep{\fill}}lrrrrrrrr@{}}
    \toprule
    & & \multicolumn{3}{c}{Denoising (s)} & \multicolumn{3}{c}{Per step (s)} & \\
    \cmidrule(lr){3-5}\cmidrule(lr){6-8}
    Model & Devices & A & B & C & A & B & C & VAE (s) \\
    \midrule
    Wan2.1-1.3B       & 1 & 147.97  & 106.29 & 81.75  & 2.960  & 2.126  & 1.635  & 2.20 \\
    Qwen-Image       & 1 & 44.37   & 42.19  & 17.20  & 0.888  & 0.846  & 0.344  & 0.09 \\
    Wan2.1-14B       & 4 & 978.60  & 716.80 & 500.76 & 19.573 & 14.335 & 10.015 & 3.13 \\
    Wan2.1-14B       & 8 & 562.94  & 406.74 & 299.26 & 11.259 & 8.135  & 5.985  & 1.87 \\
    HunyuanVideo-13B & 4 & 1002.18 & 648.67 & 540.08 & 20.044 & 12.972 & 10.801 & 11.00 \\
    HunyuanVideo-13B & 8 & 714.71  & 455.82 & 389.79 & 14.295 & 9.116  & 7.795  & 7.09 \\
    Wan2.2-A14B$^\dagger$ & 4 & 793.80 & 585.25 & 400.57 & 19.845 & 14.631 & 10.014 & 4.31 \\
    Wan2.2-A14B$^\dagger$ & 8 & 477.64 & 352.54 & 239.49 & 11.941 & 8.813  & 5.987  & 2.61 \\
    MiniMax-H3$^\dagger$  & 4 & 48.00  & 39.75  & 23.35  & 5.992  & 4.960  & 2.917  & 3.25 \\
    MiniMax-H3$^\dagger$  & 8 & 29.00  & 25.56  & 13.91  & 3.615  & 3.185  & 1.736  & 1.98 \\
    \bottomrule
  \end{tabular*}
  \par\medskip
  \begin{minipage}{0.62\textwidth}
  \caption{End-to-end scaling of BF16 from four to eight devices.}
  \label{tab:latency-scaling}
  \centering
  \small
  \setlength{\tabcolsep}{2pt}
  \begin{tabular*}{\linewidth}{@{\extracolsep{\fill}}lrrr@{}}
    \toprule
    Model & 4 devices (s) & 8 devices (s) & Speedup \\
    \midrule
    Wan2.1-14B       & 981.74  & 564.82 & $1.74\times$ \\
    HunyuanVideo-13B & 1013.19 & 721.82 & $1.40\times$ \\
    Wan2.2-A14B      & 798.49  & 480.68 & $1.66\times$ \\
    MiniMax-H3       & 51.30   & 31.03  & $1.65\times$ \\
    \bottomrule
  \end{tabular*}
  \end{minipage}
  \par\medskip
  \caption{Median Qwen3-8B serving latency and throughput.}
  \label{tab:latency-llm}
  \centering
  \small
  \setlength{\tabcolsep}{3pt}
  \begin{tabular*}{\textwidth}{@{\extracolsep{\fill}}rcrrrrr@{}}
    \toprule
    Concurrency & Group & TTFT (ms) & TPOT (ms) & End-to-end (ms) & Output tokens/s & C/A throughput \\
    \midrule
    1  & A & 97.1   & 13.68 & 7086  & 72.2   & -- \\
    1  & C & 39.3   & 6.94  & 3584  & 142.8  & $1.98\times$ \\
    8  & A & 667.0  & 15.96 & 8823  & 463.5  & -- \\
    8  & C & 243.2  & 8.37  & 4508  & 905.8  & $1.95\times$ \\
    64 & A & 3219.6 & 29.71 & 18403 & 1782.5 & -- \\
    64 & C & 1207.2 & 19.49 & 11149 & 2942.4 & $1.65\times$ \\
    \bottomrule
  \end{tabular*}
\end{table*}

\end{document}